\documentclass[sn-mathphys]{uom-cs}% Math and Physical Sciences Reference Style
\usepackage{multicol}
\jyear{2023}%
\theoremstyle{thmstyleone}%
\newtheorem{proposition}{Proposition}[section] % Define proposition environment with section dependency

\makeatletter
\@addtoreset{proposition}{section}
\makeatother

\theoremstyle{thmstyletwo}%

\theoremstyle{thmstylethree}%
\usepackage{amsmath}
\usepackage{amsthm}
\usepackage{chngcntr}

\newenvironment{boldproof}[1][\proofname]{%
  \proof[\textbf{#1}]%
}{\endproof}

\begin{document}
\title[Goal Exploration Augmentation via Pre-trained Skills]{Goal Exploration Augmentation via Pre-trained Skills for Sparse-Reward Long-Horizon Goal-Conditioned Reinforcement Learning}

%%=============================================================%%
%% Prefix   -> \pfx{Dr}
%% GivenName    -> \fnm{Joergen W.}
%% Particle -> \spfx{van der} -> surname prefix
%% FamilyName   -> \sur{Ploeg}
%% Suffix   -> \sfx{IV}
%% NatureName   -> \tanm{Poet Laureate} -> Title after name
%% Degrees  -> \dgr{MSc, PhD}
%% \author*[1,2]{\pfx{Dr} \fnm{Joergen W.} \spfx{van der} \sur{Ploeg} \sfx{IV} \tanm{Poet Laureate}
%%                 \dgr{MSc, PhD}}\email{iauthor@gmail.com}
%%=============================================================%%

\author{\fnm{Lisheng} \sur{Wu}}\email{Lisheng.Wu@manchester.ac.uk}

\author*{\fnm{Ke} \sur{Chen}*}\email{Ke.Chen@manchester.ac.uk}
% \equalcont{These authors contributed equally to this work.}

% \author[1,2]{\fnm{Third} \sur{Author}}\email{iiiauthor@gmail.com}
% \equalcont{These authors contributed equally to this work.}

\affil{\orgdiv{Department of Computer Science}, \orgname{University of Manchester}, \orgaddress{\city{Manchester}, \postcode{M13 9PL}, \country{United Kingdom}}}

%%==================================%%
%% sample for unstructured abstract %%
%%==================================%%

\abstract{Reinforcement learning (RL) often struggles to accomplish a sparse-reward long-horizon task in a complex environment. Goal-conditioned reinforcement learning (GCRL) has been employed to tackle this difficult problem via a curriculum of easy-to-reach sub-goals. In GCRL, exploring novel sub-goals is essential for the agent to ultimately find the pathway to the desired goal.  How to explore novel sub-goals efficiently is one of the most challenging issues in GCRL. Several goal exploration methods have been proposed to address this issue but still struggle to find the desired goals efficiently.
In this paper, we propose a novel learning objective by optimizing the entropy of both achieved and new goals to be explored for more efficient goal exploration in sub-goal selection based GCRL. To optimize this objective, we first explore and exploit the frequently occurring goal-transition patterns mined in the environments similar to the current task to compose skills via skill learning. Then, the pre-trained skills are applied in goal exploration with theoretical justification. Evaluation on a variety of spare-reward long-horizon benchmark tasks suggests that incorporating our method into several state-of-the-art GCRL baselines significantly boosts their exploration efficiency while improving or maintaining their performance.}


%%\pacs[JEL Classification]{D8, H51}

%%\pacs[MSC Classification]{35A01, 65L10, 65L12, 65L20, 65L70}

\maketitle

\section{Introduction}\label{sec:introduction}
Reinforcement learning (RL) has successfully solved some complex problems, e.g., board games \cite{silver2017mastering}, protein prediction \cite{jumper2021highly} and robotic locomotion tasks \cite{levine2016end}, where rewards as supervision signals play a crucial role in the learning process. Generally, it is possible to solve most if not all tasks via RL as long as the rewards are designed properly \cite{silver2021reward}. In contrast to non-trivial reward design principles, setting valuable rewards only for states that reach the desired goals is easier and can generalize across different tasks. Those tasks therefore can be easily framed as goal-conditioned reinforcement learning (GCRL) problems to target at reaching the desired goals. However, the simple reward design also makes it extremely hard for RL to learn how to reach the goals as it is hard for the agent to explore them to obtain valuable rewards for learning. The problems have become more severe in long-horizon tasks where the goals are only reachable beyond a long-horizon. Thus, under the sparse-reward design, how to explore the goals efficiently in long-horizon tasks remains a key problem for the wider applications of RL.

In sparse-reward long-horizon GCRL tasks, instead of directly targeting at the desired goals, the agent often learns to reach an implicit curriculum of sub-goals that are easier to reach and help the agent to discover the pathway to the desired goals. Following the curriculum, the agent gradually expands its reachable sub-goals to cover the desired goals. In the process, the efficiency of exploring new sub-goals for the agent to learn is essential for discovering the desired goals efficiently. Several strategies have been proposed to explore new sub-goals efficiently \cite{florensa2018automatic,pong2019skew,pitis2020maximum,mendonca2021discovering,liuetal2022gcrl-survey}. However, there still exists a large gap to the level of efficiency required by  wider RL applications.

The efficient exploration of human beings often establishes on various patterns in the interactions with the environment. Even a baby would master how to explore the room more efficiently via crawling, a kind of behavior patterns that enables the baby to move to nearby positions.
%The crawling is a kind of behavior patterns, and similar gestures of crawling often lead to similar moves in the room. Inversely, to make an expected move, the baby should adopt specific gestures. If semantically the same transitions in the goal space can be realized by the same behavior patterns, they together form goal-transition patterns.
%%%%%%%%%%%%%%%%%%%%%%%%%%%%%%%%%%%%%%%%%%%%%%%%%%%%%%%%%%%%%%%%%%%%%%%%%%%%%%
%In contrast to crawling, casually stretching its body at an earlier age can not.
We hypothesize that a key component for efficient goal exploration is to utilize the behavior patterns of the agent transitioned to goals nearby, like the baby crawling. %Such behavior patterns along with the patterns of goal transitions form \emph{goal-transition patterns}.
However, existing GCRL strategies do not take such kind of patterns into consideration. In our work, we learn such kind of behavior patterns in the form of skills \cite{florensa2017stochastic,eysenbach2018diversity} that are pre-trained on the environments of the properties shared by downstream tasks. Each skill corresponds to an individual policy for the agent to conduct specific behavior patterns. The agent is trained in the pre-training environments to visit a set of different nearby goals following each skill and those skills are transferred to downstream tasks for more efficient exploration. From the viewpoint of exploration, we are interested in behavior patterns that visit goals as widely as possible as it tends to discover more novel goals. Thus, we propose a maximum entropy objective on the distribution of achieved goals induced by following those skills.

Our main contributions are summarized as follows:
1)  We propose a maximum entropy goal exploration method, \textit{goal exploration augmentation via pre-trained skills} (GEAPS), to augment exploration in GCRL.
%2) We show that our method improves a tighter lower bound of the entropy of achieved sub-goals.
2) We introduce the entropy of goals in skill learning, which stabilizes skill learning and helps the agent gain more efficiency in goal exploration on challenging downstream tasks. Furthermore, we conduct a theoretical analysis of this entropy-based skill learning method.
3) We provide theoretical analyses for the benefits of utilizing pre-trained skills and the effectiveness achieved through our exploration strategy under specific conditions.
4) We demonstrate that incorporating our GEAPS algorithm into the state-of-the-art GCRL methods boosts their exploration efficiency for several spare-reward long-horizon benchmark tasks.

\section{Related Work}
\label{sect:related}
\normalbaroutside
\textbf{Exploration for New Goals.}~ Using uniformly sampled actions, like $\epsilon$-greedy algorithm, and introducing noises to policy actions are common strategies for exploration in RL. However, they are not sufficient to solve sparse-reward and long-horizon tasks. Different goal exploration methods have been proposed to accomplish those challenging tasks. A class of methods focus on a sub-goal selection strategy that helps with better goal exploration. Skew-Fit \cite{pong2019skew} samples sub-goals from a skewed distribution that is approximately uniform over historical achieved goals, and OMEGA \cite{pitis2020maximum} selects sub-goals by maximizing the entropy of achieved goals from low-density regions.
% HESS \cite{li2021active} and SFL \cite{hoang2021successor} discretize the goal space and prioritize the goals of lower visit counts or lower cumulative future visit counts.
Goal GAN \cite{florensa2018automatic} and the AMIGO \cite{campero2020learning} select sub-goals of intermediate difficulties that prevent the agent from getting trapped in too easy tasks and avoiding too difficult ones. By and large, however, such methods still rely on uniformly sampled actions and action noise to find new goals while pursuing sub-goals, which restricts the goal exploration to neighboring states along the trajectory to the sub-goal. To overcome this limitation, \cite{pitis2020maximum,hoang2021successor,hartikainen2019dynamical} additionally explore goals via random actions after reaching the specific sub-goal, which gives the agent larger freedom to explore around the sub-goal. Nevertheless, random actions do not involve any learned knowledge about tasks other than the action space, which restricts them from exploring a wide range of goals. In contrast, we involve behavior patterns transitioned to nearby goals in the form of skills pre-trained in similar tasks. The pre-trained skills enable transition to nearby goals quicker so that a wider range of goals can be explored within the same time steps. As a model-based method, LEXA \cite{mendonca2021discovering} trains an exploration policy in a world model of the environment to discover novel goals and perform exploration via the trained exploration policy in the environment. However, a notable lack of experiences around the novel goals makes the simulated dynamics inaccurate around them. The inaccurate simulated dynamics also prevent the exploration policy from exploring a wider range of novel goals. As a model-free method, our method does not rely on the exact dynamics around the novel goals. Instead, we explore new goals with the behavior patterns transitioned to nearby goals to increase the chance for the agent to reach the nearby goals faster than those methods without any knowledge of goal transition.\\

\noindent
\textbf{Skill Learning.} ~ To learn the behavioral patterns transitioned to nearby goals, we perform skill learning in pre-training environments with each skill learning to reach a different set of goals. To achieve this, a well-known idea is to maximize the mutual information between skills and the goals that are going to be visited, which can be expressed as follows:
\begin{subequations}
\begin{align}
    I(\mathcal{G}; \mathcal{Z}) &= H(\mathcal{Z}) - H(\mathcal{Z}|\mathcal{G}) \label{eq:reverse}  \\
                                &=  H(\mathcal{G}) - H(\mathcal{G}|\mathcal{Z}) \label{eq:forward}
\end{align}
%\label{eq:mutural-information}
\end{subequations}
% \begin{align}
%     I(\mathcal{G}; \mathcal{Z}) &= H(\mathcal{Z}) - H(\mathcal{Z}|\mathcal{G})  \label{eq:reverse} \\
%                                 &=  H(\mathcal{G}) - H(\mathcal{G}|\mathcal{Z}) \label{eq:forward}
% \end{align}
where $\mathcal{G}$ is the goal space and $\mathcal{Z}$ denotes the latent space of the skill policy where each skill is represented by the skill policy conditioned on an individual latent vector. As the state itself can be considered as a goal, we would review the related works below in terms of goals for simplicity. With Eq.~\ref{eq:reverse}, SNN4HRL \cite{florensa2017stochastic} and DIYAN \cite{eysenbach2018diversity} learn skills by fixing the distribution of latent vectors and minimizing the conditional entropy $H(\mathcal{Z}|\mathcal{G})$. DADS \cite{sharma2019dynamics} estimates $H(\mathcal{G})$ and $H(\mathcal{G}|\mathcal{Z})$ with the help of a skill-dynamics model and learns the skills via  Eq.~\ref{eq:forward}. However, their learned skills can cover only a small portion of reachable goals due to the fact that mutual information may have many optima and covering more goals does not necessarily contribute to higher mutual information, as shown in Section~\ref{subsec:skill}.
%To cover more goals, LSD \cite{park2021lipschitz} regularizes the skill learning via maximizing the alignment between the latent vector and the changes in achieved goal representations to learns far-reaching skills. \textbf{(We do not compare with it, so we can not criticize. Will remove it.)}
EDL \cite{campos2020explore} explores the goal space at first, then encode those goals into discrete latent vectors $\mathcal{Z}$ via a trained VQ-VAE \cite{van2017neural}, and finally learn each skill from the rewards based on the likelihoods of the achieved goals that are predicted by the VQ-VAE decoder. Though the skills learned via EDL can reach goals further away, they are not optimized to reach all reachable goals.  As the pre-training environments do not reveal the exact structures of downstream tasks, some behavioral patterns transitioned to nearby goals may not work out as expected. Thus, we expect the learned behavioral patterns to support as many transitions to nearby goals as possible so that they can be more robust to different situations in downstream tasks. To achieve this, we introduce an alternative objective for skill learning based on mutual information maximization. The maximized entropy of goals ensures that skills can reach a wider range of nearby goals and avoid bad local optima in goal covering. as demonstrated in our experiments reported in Section~\ref{subsect:skill-comparison}.

\section{Preliminary}
\label{sect:preliminry}
 While traditional reinforcement learning is often modeled as a \emph{Markov decision process} (MDP), GCRL augments the MDP with a goal state to form \emph{goal-augmented} MDP (GA-MDP)  \cite{schaul2015universal}. A GA-MDP $\mathcal{M}^\mathcal{G}$ is denoted by a tuple $(\mathcal{S}, \mathcal{A}, \mathcal{T}, \mathcal{G}, r, \gamma, \phi, p_{dg}, T)$ where $\mathcal{S}, \mathcal{A}, \gamma, T$ are state space, action space, discount factor and the horizon, respectively. $\mathcal{T}\!\!: \mathcal{S}\times\mathcal{A}\times\mathcal{S} \rightarrow [0, 1]$ is the transition function, $\mathcal{G}$ is the goal space, $p_{dg}$ is the desired goal distribution and $\phi: \mathcal{S} \rightarrow \mathcal{G}$ is a tractable mapping function that maps a state to its corresponding achieved goal. The reward function $r: \mathcal{S}\times\mathcal{G}\times\mathcal{A} \rightarrow \mathbb{R}$ provides the learning signals for the agent, but valuable rewards can only be obtained when the agent reaches the desired goals in the sparse-reward setting. GCRL requires the agent to learn a policy $\pi: \mathcal{S} \times \mathcal{G} \times \mathcal{A} \rightarrow [0, 1]$ to maximize the expected cumulative return:
\begin{align*}
    J(\pi) = \mathbb{E}_{\substack{g\sim p_{d\!g}, a_t \sim \pi(\cdot|s_t, g) \\ s_{t+1}\sim \mathcal{T}(\cdot|s_t, a_t)}}[\sum_{t=1}^T \gamma^t r(s_t, a _t, g)].
\end{align*}

\begin{figure*}[t]
    \centering
    \includegraphics[width=1.0\linewidth]{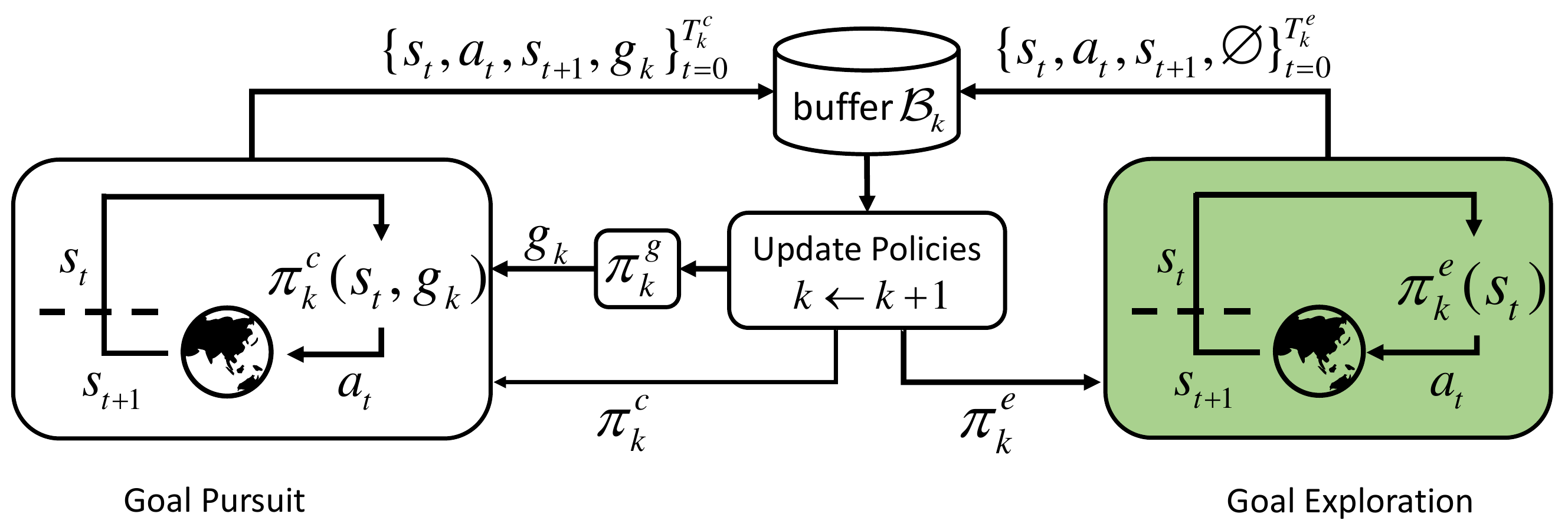}
    \caption{A generic goal-conditioned reinforcement learning (GCRL) framework for long-horizon and sparse-reward tasks.}
    \label{fig:gcrl_scheme}
\end{figure*}

In GCRL\footnote{Some GCRL methods may not have all the components in the generic framework shown in Figure~\ref{fig:gcrl_scheme}.}, the agent makes actions either in pursuit of a goal or trying to explore more goals. As depicted in Figure~\ref{fig:gcrl_scheme}, we divide the entire interaction process in an iteration into \emph{goal pursuit} and \emph{goal exploration}, depending on whether the decision policies are conditioned on goals. During policy training in the $k\text{th}$ iteration,  let $\pi_k^g$, $\pi_k^c$ and $\pi^e_k$ denote the policy deciding a behavioral goal for an agent to pursue,  the goal-conditioned policy for the agent to achieve a goal and the exploration policy for the agent to explore new goals, respectively. Before goal pursuit, a goal $g_k$ will be sampled from $\pi_k^g(\mathcal{G})$, $g_k\sim\pi_k^g(\mathcal{G})$. During goal pursuit, the agent takes an action $a_t\sim\pi_k^c(s_t, g_k)$ at each time step $t$ until $T_k^c\leq T$. For clarity, $T_k^c$ refers to the number of steps required to reach the goal $g_k$ at iteration $k$ in goal pursuit. In the goal exploration process, the agent takes actions $a_t \sim \pi_k^e(s_t)$ until $T_k^e \leq T$. To make the best use of interaction steps, we perform goal exploration subsequently after the agent achieves the goal during goal-pursuit \cite{pitis2020maximum,hoang2021successor,hartikainen2019dynamical}  instead of conducting goal exploration separately. Thus, the total steps in iteration $k$ is $T=T_k^c + T_k^e$, meaning that the number of steps taken for goal exploration, $T_k^e$, depend on $T_k^c$  in iteration $k$. The data collected from both goal pursuit and exploration are stored in the replay buffer $\mathcal{B}_k$ at iteration $k$.  In the $(k\!+\!1){\text{th}}$ iteration, $\pi_k^g,\pi_k^c,\pi_k^e$ would be updated to $\pi_{k+1}^g,\pi_{k+1}^c,\pi_{k+1}^e$ , respectively, based on the training data in the current replay buffer $\mathcal{B}_k$. Then, the updated policies will be used in the new round of data collection. Furthermore, we denote the achieved goals as the set $\mathcal{G}_{\mathcal{B}}: \{\phi(s)|s\in\mathcal{B}_k\}$ , their distribution in the goal space $\mathcal{G}$ as $p_{ag,k}(\mathcal{G})$ and their entropy as $H_{ag,k}(\mathcal{G})$ at iteration $k$. To simplify the presentation, we shall drop off the explicit iteration index, $k$, from the subscript of the above notation in the rest of the paper.

\section{Method}
\label{Sect:method}
In this section, we propose a new learning objective for goal exploration, then present skill learning via the goal-transition patterns to optimize our learning objective, which leads to our GEAPS algorithm.

\subsection{Learning Objective for Goal Exploration}
\label{sec:max_entropy_ge}
Unlike the previous works reviewed in Section~\ref{sect:related},
we focus on the goal exploration associated with goal-independent behavior. As it is hard to directly explore desired goals in long-horizon and sparse-reward tasks, a well-known learning objective is to maximize the entropy of historical achieved goal $H(\mathcal{G})$. OMEGA  \cite{pitis2020maximum} has shown how to optimize the entropy of already achieved goals, $H_{ag}(\mathcal{G})$, in the goal pursuit process. We make a step forward by analyzing how to further optimize  $H_{ag}(\mathcal{G})$ via goal exploration immediately after goal pursuit in each trial.
Let  $p_e(\mathcal{G})$ and $H_e(\mathcal{G})$ denote the distribution of goals encountered in goal exploration and its entropy, respectively.
In the goal exploration process starting with the initial state $s_0$\footnote{To simplify the notation, we designate the state, $s_{T^c}$, reached by the goal pursuit after $T^c$ transitions as the initial state, $s_0$, that triggers the goal exploration process (see Section~\ref{sect:preliminry} and Figure~\ref{fig:gcrl_scheme} for clarity).} and going through $T^e$ transitions,
we have
\begin{equation}
   p_e(g|s_0) = \frac{1}{T^e} \mathbb{E}_{\tau \sim \Pi^e(\tau)}\left[ \sum_{i=1}^{T^e} \mathbb{I}(\phi(s_{i}) = g) \right],
   \label{eq:goal_distribution}
\end{equation}
where \( \tau = (s_0, a_0, \ldots, s_{T^e}) \) denotes a trajectory that adheres to the distribution $ \Pi^e(\tau) = \prod_{i=0}^{T^e-1} \mathcal{T}(s_{i+1}|s_i, a_i)\pi^e(a_i|s_i) $ under the exploration policy \( \pi^e \). The indicator function \( \mathbb{I}(\phi(s_i) = g) \) indicates whether the state \( s_i \) achieves the goal \( g \).
After goal exploration (c.f. Figure~\ref{fig:gcrl_scheme}), the updated distribution of achieved goals  $p_{ag}'(\mathcal{G})$ is a weighted mixture of $p_{ag}(\mathcal{G})$ and $p_e(\mathcal{G})$ as follows:
\begin{equation}
    p_{ag}'(\mathcal{G}) = c\text{ } p_{ag}(\mathcal{G}) + (1-c)p_e(\mathcal{G}), \nonumber
\end{equation}
where  $c=\frac {|\mathcal{B}|+T^c} {|\mathcal{B}|+T^c+ T^e}$ and $|\mathcal{B}|$ is the size of the current replay
buffer.

To develop our learning objective for goal exploration augmentation, we formulate a proposition as follows: 
% \vspace*{-2mm}
\begin{proposition}
Let $H_{ag}'(\mathcal{G})$ represent the updated entropy of achieved goals following the goal exploration. This entropy is bounded from below by the sum of the weighted entropies of the original achieved goals and the goals encountered during goal exploration, namely, $c~H_{ag}(\mathcal{G})$ and $(1-c)~H_e(\mathcal{G})$. That is,
\begin{align}
    H_{ag}'(\mathcal{G}) &\geq cH_{ag}(\mathcal{G}) + (1-c)H_e(\mathcal{G}). \label{eq:lower-bound}
\end{align}
\label{math:p1}
\end{proposition}
% \vspace*{-5mm}
The proof of Proposition \ref{math:p1} can be found in Appendix \ref{appendix-0}.
According to Eq.~(\ref{eq:lower-bound}), an increase in $H_{ag}(\mathcal{G})$ and $H_e(\mathcal{G})$ elevates the lower bound of the resulting entropy $H_{ag}'(\mathcal{G})$. As  the OMEGA \cite{pitis2020maximum} asserts, $H_{ag}(\mathcal{G})$ can be maximized by selecting low-density goals as sub-goals. However, optimizing $H_e(\mathcal{G})$ is challenging due to the agent's limited understanding of new sub-goal dynamics, which may necessitate arbitrary exploration. 

Despite unknown dynamics, we observe that overlapping elements may exist between the agent's transition mechanisms and a pre-training environment. These shared features form goal-transition patterns, beneficial for exploring unfamiliar goals. When all goal-transition patterns are available in a new sub-goal, the generic entropy of explored goals $H_e(\mathcal{G})$ is denoted by $\hat{H}_e(\mathcal{G})$.
To optimize $\hat{H}_e(\mathcal{G})$, an exploration policy must aim to visit as many goals as feasible within a given time frame, while avoiding revisits and maintaining stochasticity. Backed by  theoretical justification presented in Section~\ref{Sect:ta-1-1}, we suggest developing an exploration policy based on an array of stochastic pre-trained skills. Each skill targets a maximum set of sub-goals, leading to a maximized $\hat{H}_e(\mathcal{G})$. Although this assumption may not apply during actual exploration, $\hat{H}_e(\mathcal{G})$ still acts as an upper bound of $H_e(\mathcal{G})$ even though missing goal-transition patterns lead to failed transitions. Hence, enhancing $\hat{H}_e(\mathcal{G})$ could significantly improve the agent's exploration efficiency.

\subsection{Skill Acquisition}
\label{subsec:skill}

 For a given environment, optimizing our learning objective in  Eq.~\ref{eq:lower-bound} leads to the maximum entropy of goals to be explored in goal exploration   (c.f. Figure~\ref{fig:gcrl_scheme}). However, the exact dynamics around the current state is often unknown, hence it is infeasible to directly maximize the entropy of goals to be explored via  $p_e(\mathcal{G})$ in  Eq.~\ref{eq:goal_distribution}. Fortunately, this issue can be addressed with the auxiliary  information named goal-transition patterns. A goal transition always has a starting goal $g_s$ and an end goal $g_e$ but goal transitions of the same $g_s$ and $g_e$ may involve different intermediate states. Here, we define a \textit{goal-transition pattern} as a goal transition process that can transit across different states with actions  but preserves the same properties independent of $g_s$ and $g_e$ in the goal space $\mathcal{G}$. It is analogous to image recognition where an object's identity is independent of its location in the image.
Exploring with a goal-transition pattern from a state tends to make the changes specified by the pattern via goal-independent actions in $\mathcal{G}$. Goal-transition patterns enable planning in $\mathcal{G}$ to avoid the canceling-out effect of different actions used for goal exploration.
Composing a set of frequently occurring inherent goal-transition patterns, named \emph{skills}, in a manner that maximizes the entropy of goals to be explored enables an agent to expand its achieved goal space more efficiently for better goal covering. Such skills can be learned via another policy as described below.

Although we cannot find all the frequently occurring goal-transition patterns without traversing the entire environment, we observe that there are many goal-transition patterns in common that can be mined from similar environments via pre-training. A pre-training environment should share both the same agent space $\mathcal{S}^{agent}$ \cite{florensa2017stochastic, konidaris2007building} and the same goal space $\mathcal{G}$ with the current task. The agent space $\mathcal{S}^{agent}$ is simply a shared subspace of the state space $\mathcal{S}$ and semantically the same across a collection of relevant tasks. $\mathcal{S}^{agent}$ generally does not convey goal information since the transition dynamics in their goal spaces often differ on the pre-training tasks. In our work, $\mathcal{S}^{agent}$ needs to be independent of the goal space $\mathcal{G}$  of any tasks. Thus, the goal-transition patterns can be transferred to a GCRL task within $\mathcal{S}^{agent}$ via learned policies that execute the inherent goal-transition patterns mined in the pre-training environments.
As our ultimate goal is to learn the composition of goal-transition patterns or skills, we can directly learn another policy that maximizes the expected entropy of goals to be explored in the  pre-training environments without modeling the behavior for each goal-transition pattern explicitly. Thus, the behavior of frequently occurring goal-transition patterns is automatically encoded by the policy via learning. We formulate such policy learning as a \textit{skill learning} process.

\subsection{Skill Learning}
\label{subsect:skill-learn}

We denote a skill by a latent vector $\pmb z$, the set of all the pre-trained skills by $\mathcal{Z}$, and the corresponding multi-modal skill policy by $\pi_\mathcal{Z}$. For each skill, $\pi_\mathcal{Z}$ would select an action $a_t \sim \pi_\mathcal{Z}(a_t|s^{agent}_t, \pmb z)$. To learn a set of diverse skills, we formulate its learning objective as the mutual information between the skills and the goals conditioned on initial goal states by Eqs.~\ref{eq:reverse} and \ref{eq:forward}. However, previous skill learning methods often fail to learn a wide coverage of goals, which is attributed to the fact that there exist many optima in the mutual information function and covering more goals does not always lead to higher mutual information. Without loss of generality, we assume both the goal space $\mathcal{G}$ and the latent space $\mathcal{Z}$ for skills are discrete and it is common to have $|\mathcal{G}|>|\mathcal{Z}|$. 
% As illustrated in Figure~\ref{fig:plot-goal-entropy}, 
Even when the mutual information $I(\mathcal{G}; \mathcal{Z})$ has been maximized to be $\text{log}|\mathcal{Z}|$ via  Eq.~\ref{eq:reverse}, the entropy of goals $H(\mathcal{G})$ can still vary from $\text{log}|\mathcal{Z}|$ to $\text{log}|\mathcal{G}|$. When $H(\mathcal{G})$ takes low values, the goal coverage appears poor, which motivates us to develop an alternative skill learning strategy.

Unlike the prior skill learning works, e.g., SNN4HRL  \cite{florensa2017stochastic} and DIAYN  \cite{eysenbach2018diversity},
we want a diverse set of skills by maximizing both $I(\mathcal{Z}, \mathcal{G})$ and $H(\mathcal{G})$. In our work, we do not maximize $H(\mathcal{G})$ directly but $H(\mathcal{G|Z})$ instead given the fact that when $I(\mathcal{Z}, \mathcal{G})$ is maximized,  Eq.~\ref{eq:forward} leads to
\begin{align}
    H(\mathcal{G}) =& I(\mathcal{Z}, \mathcal{G}) + H(\mathcal{G} | \mathcal{Z}) \nonumber\\
    = &H(\mathcal{Z}) - H(\mathcal{Z}|\mathcal{G}) + H(\mathcal{G} | \mathcal{Z})\nonumber\\
    % =& E_{s_0, z,  g}[\text{log }p(z| g, s_0) - \text{log }p(z|s_0)-\text{log }p( g|z, s_0)]. \label{eq:goal_expansion_skills}
    =& E_{\pmb z,  g}[\text{log }p(\pmb z| g) - \text{log }p(\pmb z)-\text{log }p( g|\pmb z)]. \label{eq:goal_expansion_skills}
\end{align}

In the skill learning process, however, we still cannot obtain the exact $p(\pmb z| g)$ and $p(g|\pmb z)$ that requires integration over all reachable goals and skills. We approximate $p(\pmb z| g)$ and $p(g|\pmb z)$ with $q(\pmb z| g)$ and $q(g|\pmb z)$ by using the Monte Carlo method. Motivated by the previous works  \cite{florensa2017stochastic,eysenbach2018diversity}, we set the reward for mutual information maximization as
\begin{equation}
    r^I_{z}(s_t, a_t) = \text{log }q(\pmb z| \phi(s_{t})) - \text{log }p(\pmb z). \label{reward:mutual_information}
\end{equation}
To maximize the entropy of $H(\mathcal{G}|\mathcal{Z})$, the distribution $p(g|\pmb z)$ is expected to be as uniform as possible. Thus, we design the reward as follows:
\begin{equation}
    r^H_{z}(s_t, a_t) = \text{log} \left[\text{max}_{\hat{g}} q(\hat{g}|\pmb z) - q( \phi(s_{t+1})|\pmb z)\right]. \label{reward:entropy}
\end{equation}
Here, we encourage visiting those goals less explored. Thus, it will converge when the distribution of goals to be explored are uniform. Combining the rewards specified in Eqs.~\ref{reward:mutual_information} and \ref{reward:entropy}, we achieve the pseudo reward for our skill training as follows:
\begin{equation}
    r_{z}(s_t, a_t) = r^I_{\pmb z}(s_t, a_t) + \beta r^H_{\pmb z}(s_t, a_t). \label{reward:skill}
\end{equation}
where $\beta$ is a coefficient to trade-off between $r^I_{z}$ and $r^H_{z}$.
%In our implementation, we use the TRPO \cite{schulman2015trust} in skill learning for the stability in learning (see Appendix A for details).

In GCRL, transitioning between goals $g$ and $g'$ is often represented as $g \to g + \Delta(g, g')$, where $\Delta G = \Delta(g, g')$ signifies the desired goal transition. To optimize $\hat{H}_e(\mathcal{G})$, we actually learn the skills by maximizing $H(\Delta \mathcal{G})$ in a pre-training environment.

\begin{algorithm}[t]
\caption{Goal Exploration Augmentation via Pre-trained Skills (GEAPS)}
\label{alg:geaps}
\textbf{Given:} Skill space $\mathcal{Z}$, pre-trained skill policy $\pi_\mathcal{Z}$, initial state for goal exploration $s_0$, skill horizon $T^s$, goal exploraton horizon $T^e$, replay buffer $\mathcal{B}$.
\begin{algorithmic}[1] %[1] enables line numbers
\Procedure{GEAPS}{}
% \State $i \gets 0$, $t \gets t_g$
\While{$t \leq T^e$}
\If {$t\text{ mod } T^s = 0$}
\State sample a skill $z\sim p(z|s_t)$
\EndIf
\State sample the action $a_t \sim \pi_{\mathcal Z}(s^{agent}_t, \pmb z)$
\State $s_{t+1} \sim \mathcal{T}(s_{t+1}|s_t, a_t)$
\State $t \gets t+1$

\State save $(s_t, a_t, s_{t+1}, \varnothing)$ in replay buffer $\mathcal{B}$.
\EndWhile
\EndProcedure
\end{algorithmic}
\end{algorithm}

%\paragraph{goal exploration Strategy}
\subsection{Goal Exploration Augmentation Strategy}~
The trained skill policy is used in $\pi_e$ for goal exploration (c.f. Figure~\ref{fig:gcrl_scheme}). During goal exploration, goal-transition patterns are not available in all states, hence sometimes the desired transition of skills are unreachable. Thus, we switch a skill in every $T^s$ ($T^s < T^e$) steps. We summarize our GEAPS algorithm  in Algorithm \ref{alg:geaps} that enables the trained skills $\pi_\mathcal{Z}$ to be used for learning an exploration policy $\pi_e$ during goal exploration. 

As depicted in Figure~\ref{fig:gcrl_scheme}, our GEAPS algorithm can be easily incorporated into the generic GCRL framework to improve exploration efficiency. In the goal pursuit process, an existing GCRL algorithm, e.g., Goal GAN \cite{florensa2018automatic}, Skew-Fit \cite{pong2019skew}, or OMEGA \cite{pitis2020maximum}, used in our experiments, is employed to learn a policy, $\pi_k^g$, for an agent to decide a behavioral goal to pursue and a goal-conditioned policy, $\pi_k^c$, for an agent to achieve a goal. After the $k$th round of goal pursuit is completed, our GEAPS algorithm is invoked in the goal exploration stage to learn an exploration policy, $\pi_k^e$, for the agent to explore new goals. Alternating the goal pursuit and exploration processes makes the two algorithms reach a synergy to solve a sparse-reward long-horizon reinforcement learning task.

\subsection{Theoretical Analysis}~
In this subsection, we provide theoretical analyses concerning our entropy-maximization-based methods for skill learning and goal exploration.  The proofs of those propositions can be found in Appendix \ref{appendix-0}.

\subsubsection{On the Role of Skill Composition in Learning Optimal Exploration Policy}
\label{Sect:ta-1-1}
Under the exploration policy $\pi^e$, we characterize $\Omega$ as the set of all possible trajectories encompassed within the exploration horizon $T^e$. Each exploration trajectory, represented by $\tau$, conforms to the distribution portrayed by $\tau \sim p(\Omega)$. Consequently, the entropy $\hat{H}_e({\mathcal{G}})$ can be articulated as the sum of the mutual information $I(\Omega; \mathcal{G})$ and the conditional entropy $H(\mathcal{G}|\Omega)$.  This can be represented mathematically as follows:
\begin{align}
\hat{H}_e({\mathcal{G}}) &= I(\Omega; \mathcal{G}) + H(\mathcal{G}|\Omega). 
\nonumber
%\label{eq:entropy_decomposition}
\end{align}

To maximize the mutual information $I(\Omega; \mathcal{G})$, we strive for each trajectory to cover a distinct subset of goals. With respect to optimizing $H(\mathcal{G}|\Omega)$, we aim for each trajectory to visit as many goals as feasible, maintaining uniform probability for visiting each goal within its respective subset. In exploration scenarios deploying uniform primitive actions, each trajectory bears an equivalent likelihood $\frac{1}{|\Omega|}$ of being generated. It is crucial to note that some trajectories may be limited to a single goal, whereas a specific goal could be visited by numerous trajectories.

For the optimal exploration policy, we denote the set of trajectories following the policy as $\Omega^*\subseteq\Omega$, with their corresponding distribution represented as $p_{\Omega^*}$. Consequently, this optimal exploration policy culminates in the optimal entropy of prospective goals, denoted by $\hat{H}^*_e({\mathcal{G}})$, as given as follows:
\begin{equation}
  \hat{H}_e^*({\mathcal{G}}) = I(\Omega^*; \mathcal{G}) + H(\mathcal{G}|\Omega^*). \nonumber
  %\label{eq:optimal_entropy}
\end{equation}

However, directly optimizing the exploration policy within the trajectory space $\Omega$ to obtain $\Omega^*$ and $p_{\Omega^*}$ poses significant computational challenges. This complexity primarily arises from the exponential growth in the size of $|\Omega|=|\mathcal{A}|^{T^e}$ as a function of $T^e$. This difficulty is further compounded by the potential continuity of the action space. To mitigate these challenges, we consider the prospect of simplifying the optimization problem.

\begin{proposition}
    The optimal exploration policy leading to $\Omega^*$ with the distribution $p_{\Omega^*}$ can be composed via a set of skills $\mathcal{Z}$ ($|\mathcal{Z}|<<|\Omega^*|$).
\end{proposition}

Following this proposition, we firmly advocate the prospect of pre-trained skills as a viable mechanism for acquiring optimal settings. This approach offers a captivating alternative to the direct optimization method, providing promising opportunities for  efficient exploration.

\subsubsection{On the Role of Pre-trained Skills in Improving
Exploration Efficiency}

In a given environment, the occurrence of a goal-transition $\Delta G$ can be recurrent,
and this repetitiveness is evident through various distinctive goal-transition patterns.
To systematically understand these patterns, we formulate a goal-transition pattern
as $\psi=\{s^{agent}_{start}, s^{agent}_{end}, \Delta G, \Delta A\}$.
Within this formulation, $s^{agent}_{start}$ and $s^{agent}_{end}$, which belong to the agent
state space $\mathcal{S}^{agent}$, are the initial and terminal states
of the agent for the goal-transition $\Delta G$, respectively.
$\Delta A$ denotes a sequence of actions that result in the accomplishment
of $\Delta G$. The term
$|\Delta A|$ signifies the length of the action sequence $\Delta A$, and
the cardinality of $\psi$ is denoted by $|\psi|=|\Delta A|$. The
entirety of existing goal-transition patterns is symbolized by $\Psi$.

We now delve into an analysis of how goal-transition patterns can
enhance exploration efficiency. Under the assumption that all
goal-transition patterns in the pre-training environment are accessible,
we aim to optimize $\hat{H}_e(\mathcal{G})$. Our initial focus lies on
discerning the associations between these goal-transition patterns and
individual episodes.

\begin{proposition} \label{prop:goal-transition-decomposition}
Given the horizon $T$, every trajectory $\tau$ can be decomposed into a
sequence of goal-transition patterns.
\end{proposition}

With the established connection between each trajectory and its
corresponding goal-transition patterns in Proposition
\ref{prop:goal-transition-decomposition}, we proceed to analyze the
potential of enhancing exploration efficiency based on the
goal-transition decomposition of each trajectory.

\begin{proposition} \label{prop:goal-transition-improvement}
Given an exploration horizon of $T^e$, the substitution of
goal-transition patterns within each trajectory $\tau\in\Omega$ with
alternative patterns of smaller cardinality can yield equivalent
exploration outcomes using an average number of steps that is less than
or equal to the specified $T^e$.
\end{proposition}

Proposition \ref{prop:goal-transition-improvement} presents a potential
avenue for enhancing exploration efficiency over the exploration policy
that relies on uniform primitive actions. In a practical scenario, goal-transition patterns of different cardinality coexist for the same goal transition, so the
employment of goal-transition patterns typically results in a
requirement for fewer time steps than $T^e$. Furthermore, the skill
learning methodologies described in Sections \ref{subsec:skill} and
\ref{subsect:skill-learn} are designed to exploit the potential of
goal-transition patterns, thereby further improving the exploration efficiency.

\section{Experiment}
\label{sect:experiment}

In this section, we evaluate the advantage of our GEAPS in terms of  \emph{success rate} and \emph{sampling efficiency} on a set of sparse-reward and long-horizon GCRL benchmark tasks and demonstrate the effectiveness of our pre-trained skills in our GEAPS algorithm via a comparative study.

\begin{figure}[h]
     \centering
     \includegraphics[width=0.8\linewidth]{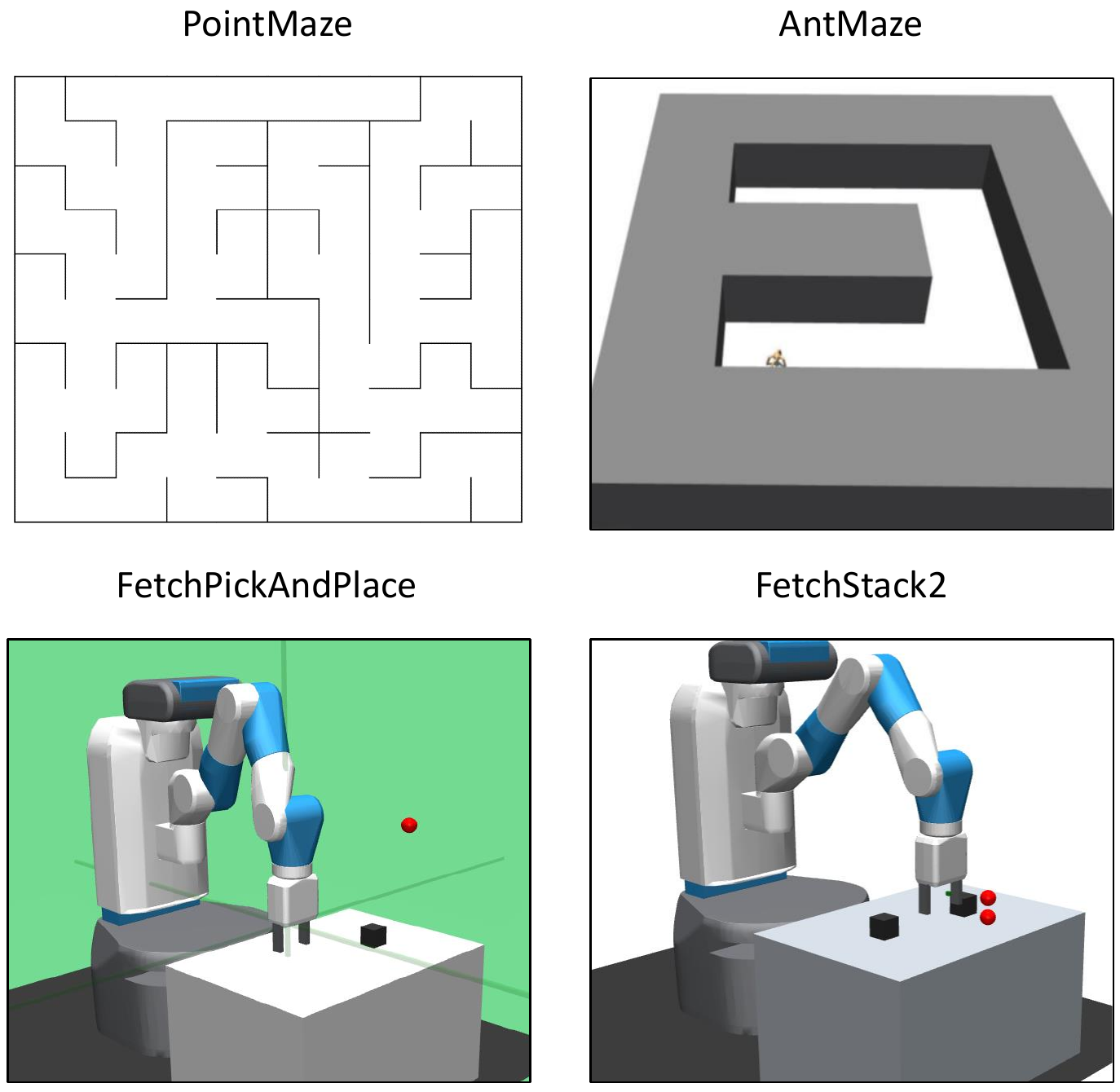}
    \caption{Four sparse-reward and long-horizon benchmark tasks.}
    \label{fig:env-vis}
%\vspace*{-5mm}

\end{figure}

\subsection{Environments and Baselines}

\subsubsection{Environments}
 As shown in Figure~\ref{fig:env-vis}, we select four common long-horizon and sparse reward environments in our experiments.
i) \texttt{PointMaze} \cite{pitis2020maximum,trott2019keeping}: a 2-D maze task that a point navigates through a $10 \!\times\! 10$ maze from the bottom left corner to the top right one. Its observation is a two-dimensional vector indicating its position in the maze. The agent may be easily trapped in somewhere with dead ends hence hardly exploring new goals.
ii) \texttt{AntMaze} \cite{pitis2020maximum,trott2019keeping}: a robotic locomotion task that controls a 3-D four-legged robot through a long U-shaped hallway to reach the desired goal position. Its thirty-dimensional observation includes the robot's status and its location. The agent can only move in a slow and jittery manner, hence it is hard to explore new goals and learn goal-reaching behavior.  
iii) \texttt{FetchPickAndPlace} (hard version) \cite{plappert2018multi}: a robot arm task where a robot arm grasps a box and moves it to a target position. The agent observes the positions of both gripper and target box as a 30-D vector, and its goal is a 3-D vector about the target position for the box. Another robot arm task, \texttt{FetchStack2} \cite{nair2018overcoming}, aims to stack two boxes at a target location, requiring the agent to move them to target positions in order. Its observation and goal are 40-D and 6-D, respectively. The agent receives no reward until placing both boxes in the correct positions, and involving two boxes makes it more difficult to explore desired goals. Reaching the desired goals once on \texttt{PointMaze} and \texttt{AntMaze} is considered as a success, while the agent is only considered to succeed on \texttt{FetchPickAndPlace} and \texttt{FetchStack2} if it still satisfies the conditions of desired goals at the end of episodes.

\subsubsection{Baselines}
%\textbf{i}) \textbf{Goal GAN} \cite{florensa2018automatic}: a typical heuristic-driven method where the intermediate difficulty is used as a heuristic to select sub-goals via a generative model.
i) \textbf{Goal GAN} \cite{florensa2018automatic}: a typical heuristic-driven method where the intermediate difficulty is used as a heuristic to select sub-goals via a generative model.
ii) \textbf{Skew-Fit} \cite{pong2019skew}: an effective exploration-based method where sub-goals are generated via sampling from a learnt skewed distribution that is approximately uniform on achieved goals.
iii) \textbf{OMEGA} \cite{pitis2020maximum}: yet another effective exploration-based method where sub-goals are generated via sampling from the low-density region of an achieved goal distribution.
While our GEAPS algorithm is applied to the above baselines to augment their goal exploration, we also compare three augmented GCRL models to the state-of-the-art (SOTA) model-based exploration in LEXA \cite{mendonca2021discovering} of which explorer can augment suitable baselines, e.g., GCSL \cite{ghosh2020learning} and DDPG \cite{lillicrap2015continuous}.

\begin{figure}[t]
     \centering
     \includegraphics[width=0.8\linewidth]{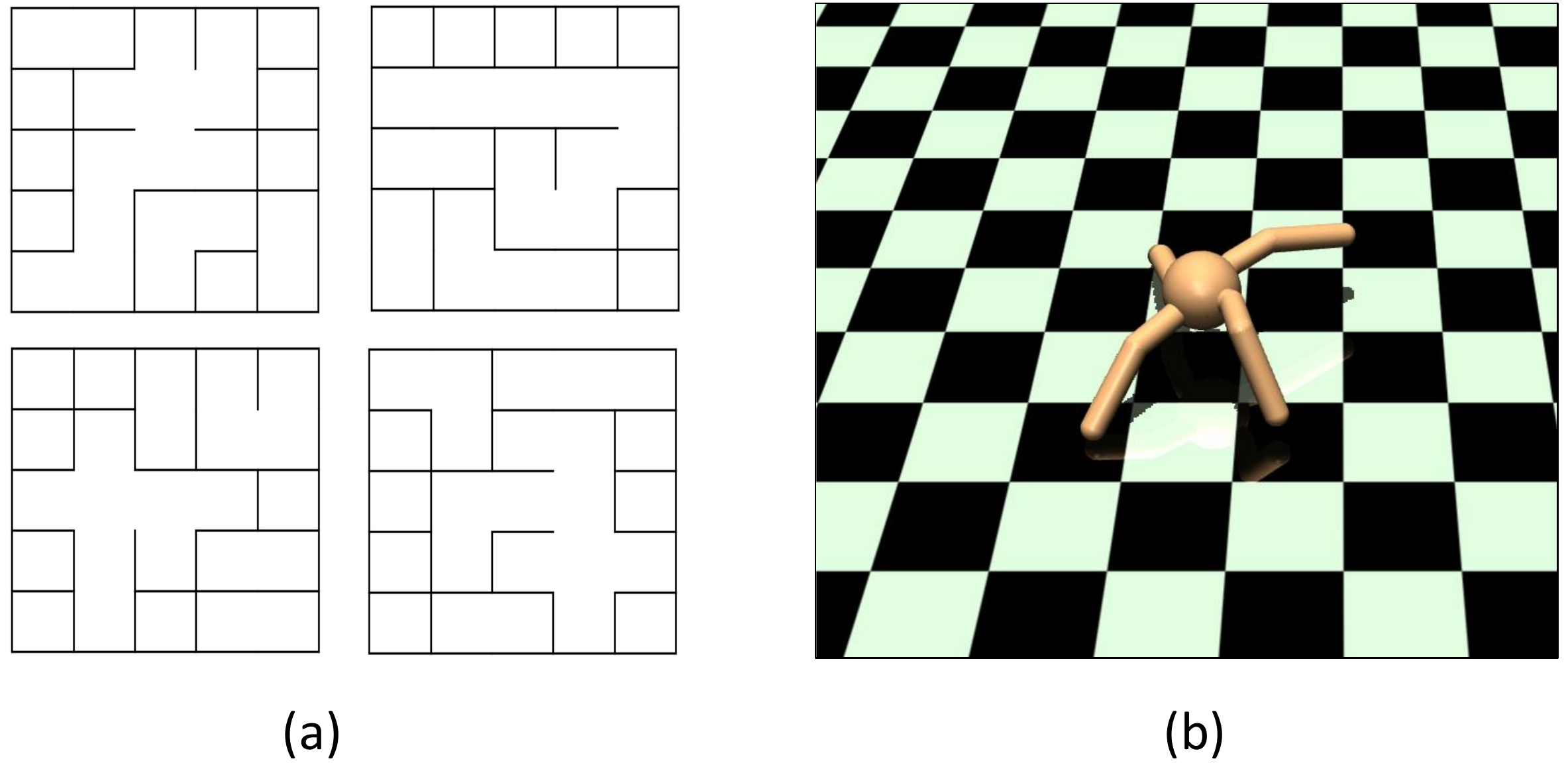}
    \caption{(a) Four typical pre-training environments for \texttt{PointMaze}. (b) The pre-training environment for \texttt{AntMaze}.}
    \label{fig:vis-pretrain-env}
%\vspace*{-5mm}
\end{figure}

\subsection{Experimental Settings and Implementation}

\subsubsection{Experimental Settings}
\label{subsect:setting}

Our experiments study the following questions:
\textbf{Q1}) How much is the sampling efficiency gained when our GEAPS is incorporated into a baseline on the condition that its performance is maintained or even improved?
\textbf{Q2}) What are the behavioral changes resulting from incorporating our GEAPS into a baseline?
\textbf{Q3}) Can our augmented  models reach the performance yielded by compared to the model-based SOTA exploration in LEXA \cite{mendonca2021discovering}?
\textbf{Q4}) What are the pre-trained skills resulting from our skill learning objective in contrast to those generated by the established skill learning methods such as SNN4HRL \cite{florensa2017stochastic} and EDL \cite{campos2020explore}?

For each baseline, we apply our GEAPS in goal exploration by keeping its original settings unchanged. Thus, we achieve three augmented models: \textbf{Goal GAN+GEAPS}, \textbf{Skew-Fit+GEAPS} and \textbf{OMEGA+GEAPS},
corresponding to three baselines. Five trials with different random seeds in each environment are conducted for reliability. Evaluation is made with a fixed budget; i.e., the training will be terminated after a (pre-set) number of steps if an agent still fails to reach the ultimate goals. The performances are evaluated in terms of the \textit{success rate}, the most important performance evaluation criterion in reinforcement learning, and the \textit{entropy of achieved goals}, a widely used evaluation criterion on sampling efficiency in GCRL.

\subsubsection{Pre-training Settings}
i) \texttt{PointMaze}: We generate 20 small $5\! \times \!5$ small mazes as pre-training environments, which include various topographies.
In each episode, the agent is initialized in a random position of the central grid.
In Figure~\ref{fig:vis-pretrain-env}(a), we exemplify four typical pre-training environments. The observation for the skill policy is the relative position with regard to the grid where the agent is located. We pre-train the skills of horizon two over those mazes via maximizing the average cumulative return averaging over the learning objective.
ii) \texttt{AntMaze}: 
We pre-train the skills on the \texttt{Ant} environment as shown
in Figure~\ref{fig:vis-pretrain-env}(b), which keeps the same 3-D four-legged robots in an open environment and sets the skill horizon as 100.
iii) \texttt{FetchPickAndPlace} and \texttt{FetchStack2}: 
The goals of two robot arm tasks are defined as the target positions of the relevant objects. Achieving a sub-goal during goal pursuit, typically inferred when the arm continues to hold the object, provides the basis for subsequent skill development. Consequently, each skill is deliberately handcrafted to direct the object along a random trajectory within a predetermined range, ensuring that collectively, the skills span all possible directions. This strategy guarantees an equal likelihood of encountering all potential goals within the boundary established by the exploration horizon $T^e$.

\subsubsection{Implementation}
Our GEAPS is implemented with the \texttt{mrl:modular RL} codebase \cite{mrl} and all the baselines adopt DDPG \cite{lillicrap2015continuous} to train goal-conditioned behavior\footnote{The code is available at:
\url{https://github.com/GEAPS/GEAPS}.}.
For three baselines
\cite{florensa2018automatic,pong2019skew,pitis2020maximum}, we use the source code provided by the authors and strictly adhere to their instructions in our experiments. LEXA is composed of a model-based exploration policy and a model-based goal-conditioned policy. As our focus is goal exploration, we use its exploration policy only and adopt model-free goal-conditioned policies optimized via GCSL \cite{ghosh2020learning} and DDPG \cite{lillicrap2015continuous} for a fair comparison. To obtain the pre-trained skills for \texttt{PointMaze} and \texttt{AntMaze}, we use a multi-layer perceptron trained with the TRPO \cite{schulman2015trust} for stability in skill learning, while the skills for two robot arm tasks are handcrafted as moving along a direction sampled uniformly in various ranges.
To pre-train the skills, we fix $\beta=0.1$ in Eq.~\ref{reward:skill} for all our experiments.
For goal exploration, we set the skill horizon $T^s$ as 2, 25, 8 and 5 for \texttt{PointMaze}, \texttt{AntMaze}, \texttt{FetchPickAndPlace} and \texttt{FetchStack2}, respectively. 

Appendixes \ref{appendix-1} and  \ref{appendix-2} describe more technical and implementation details regarding the baselines and the skill learning used in our comparative study.

\subsection{Experimental Results}

We report the main experimental results to provide the answers to four questions posed in Section~\ref{subsect:setting}.

\begin{sidewaysfigure*}
\begin{minipage}{\linewidth}
     \centering
     \includegraphics[width=\linewidth]{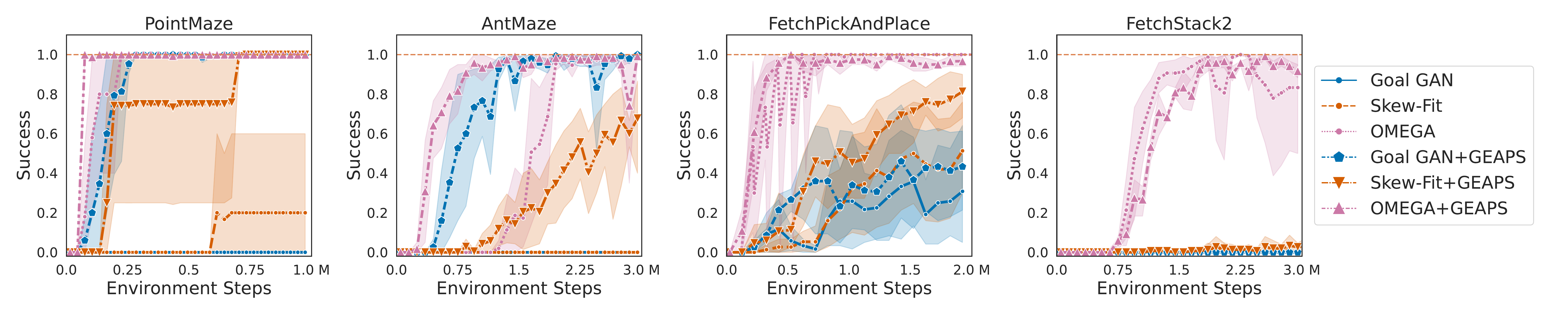}
    \caption{Test success on the desired goal distribution throughout training on four environments for the baselines and the augmented models. }
    \label{fig:vis-comparative-res}
\end{minipage}
\begin{minipage}{\linewidth}
     \vspace*{15mm}
     \centering
     \includegraphics[width=\linewidth]{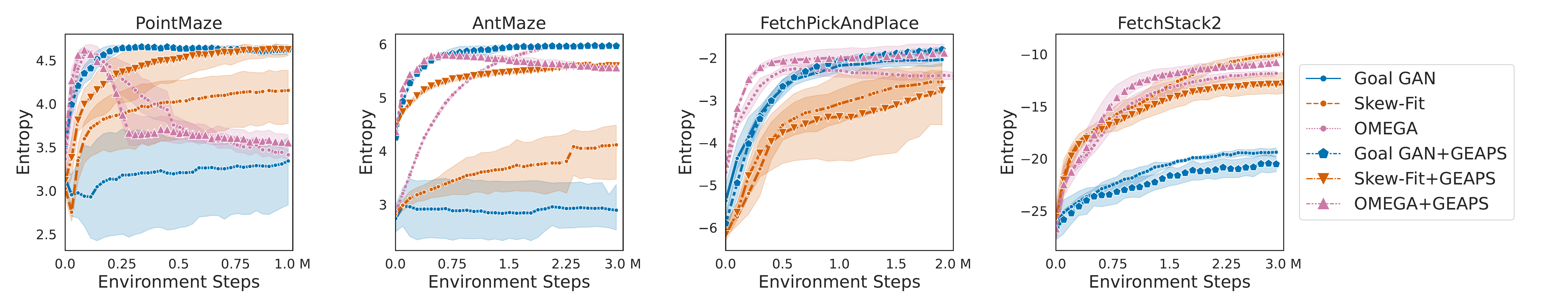}
    \caption{Empirical entropy of the achieved goal distribution throughout training on four environments for the baselines and the augmented models.}
    \label{fig:vis-entropy}
\end{minipage}
\end{sidewaysfigure*}

\subsubsection{Results on Goal Exploration Augmentation}
\label{subsect:augment}
To answer the first question, we report the results yielded by the baselines and their corresponding augmented models to gauge the gain made by our GEAPS.
In our experiments, we terminate the training at one million steps for  \texttt{PointMaze}, two million steps for \texttt{FetchPickAndPlace}, three million steps for \texttt{AntMaze} and  \texttt{FetchStack2}. An episode consists of 50 steps for \texttt{PointMaze}, \texttt{FetchPickAndPlace} and \texttt{FetchStack2} and 500 steps for \texttt{AntMaze}, respectively. We report statistics (mean and standard deviation) over five seeds in each environment.

As shown in Figure~\ref{fig:vis-comparative-res}, our GEAPS has improved three baselines in different scales across the four environments. On \texttt{PointMaze}, Goal GAN is unable to solve the environment and Skew-Fit only manages to get 20\% success at maximum. OMEGA achieves 100\% success in about 0.2 million steps. In contrast, our GEAPS enables both Goal GAN and Skew-Fit to solve \texttt{PointMaze} to achieve 100\% success in about 0.3 and 0.7 million steps. OMEGA+GEAPS is approximately twice faster as OMEGA to achieve 100\% success. On \texttt{AntMaze}, we observe similar results; both Goal GAN and Skew-Fit fail in three million steps, while our GEAPS enables Skew-Fit to solve the environment with 60\% success rates and even boost Goal GAN to have comparable results with OMEGA+GEAPS.  OMEGA+GEAPS is about three times faster than OMEGA to reach over 90\% success. On \texttt{FetchPickAndPlace}, our GEAPS boosts the success rates of Goal GAN and Skew-Fit by around 10\% percent and 20\% percent, respectively, for the same time steps. Although OMEGA+GEAPS reaches 100\% success almost at the same time as the baseline, it is 40\% faster than the baseline to reach 80\% success. On \texttt{FetchStack2}, Goal GAN, Skew-Fit along with their augmented versions hardly solve the problems with at most 7\% success observed for Skew-Fit+GEAPS and OMEGA+GEAPS yields the results comparable to OMEGA, which could be explained with the entropy of the achieved goal distribution.

As the entropy of the achieved goal distribution reflects the coverage of achieved goals in an environment, we show the empirical entropy of the achieved  goals for the baselines and the augmented models in Figure  \ref{fig:vis-entropy}.
On \texttt{PointMaze} and \texttt{AntMaze}, the entropy increases faster with the help of our GEAPS, and the improvements are especially dramatic on Goal GAN and Skew-Fit. On \texttt{FetchPickAndPlace}, GEAPS boosts the entropy of all three baselines on small scales at the beginning. On \texttt{FetchStack2}, GEAPS improves the entropy of OMEGA while deteriorating the entropy of Goal GAN and Skew-Fit marginally. The reason on the unimproved entropy on \texttt{FetchStack2} can be attributed to GEAPS that controls the goal-transition patterns for one object while making the other with little change, which leads to a slightly lower entropy of achieved goals.

We notice that in OMEGA \cite{pitis2020maximum}, several classical or SOTA baselines were tested on the same environments used in our work. As OMEGA beats those baselines with a huge margin, e.g., OMEGA is around 100 and 10 times faster than the best performer, PPO+SR \cite{trott2019keeping}, in solving \texttt{PointMaze} and \texttt{AntMaze}, respectively. Thus, the performance of OMEGA+GEAPS allows us to claim a bigger gain over those baselines used for comparison in OMEGA \cite{pitis2020maximum}.

\subsubsection{Visualization of Exploration Behavior}
\label{subsect:visual}

To answer the second question, we visualize the final achieved goals and trajectories of goal selection at the end of the episodes. The visualization vividly exhibits the behavioral changes resulting from our goal exploration augmentation.

\begin{figure}[t]
     \centering
     \includegraphics[width=1.0\linewidth]{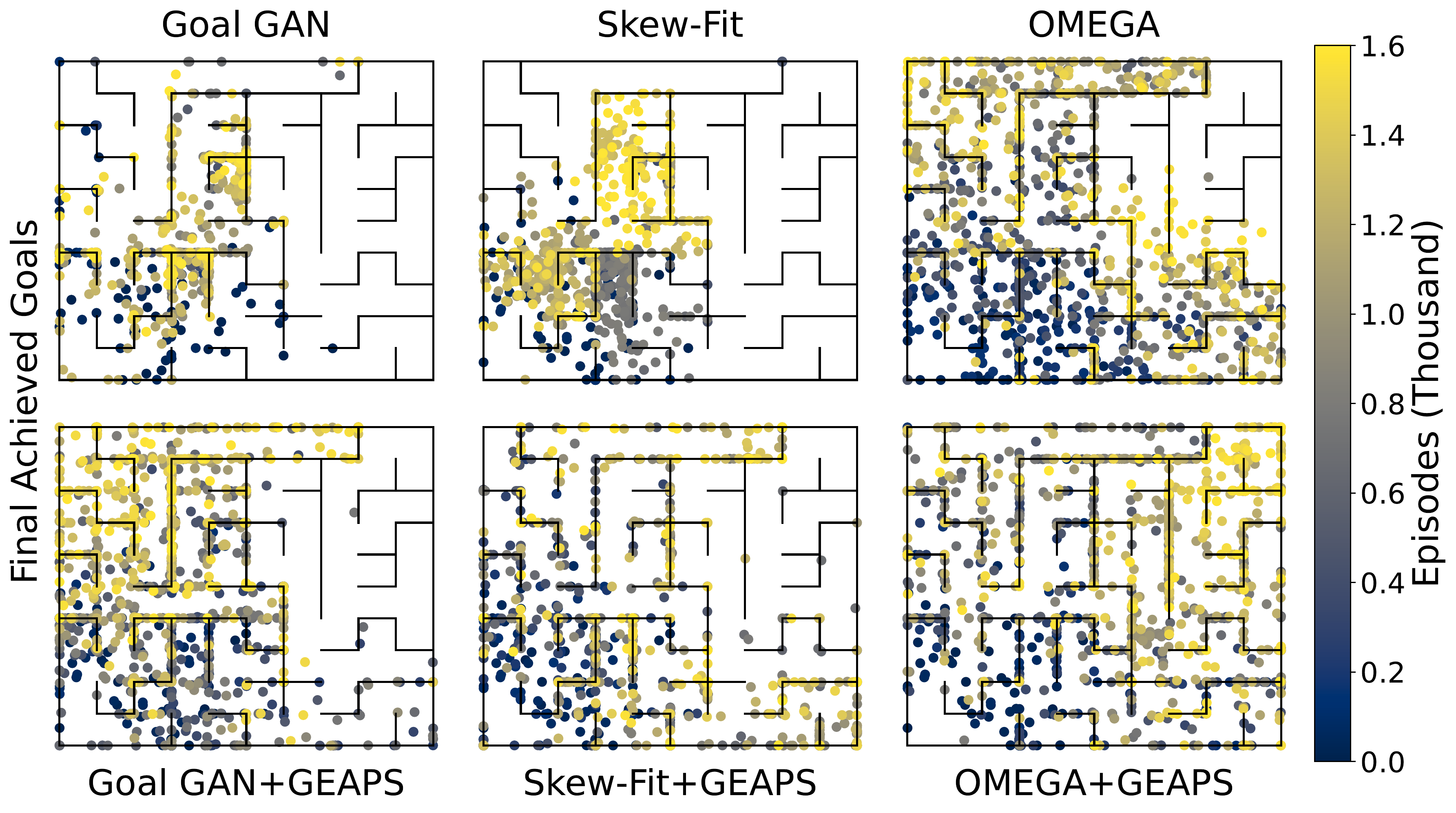}
     \vspace*{-5mm}
    \caption{Visualization of the final achieved goals  in \texttt{PointMaze}: the baselines (top) vs. the augmented models (bottom), where the training evolution process is indicated with the heatmap. }
    \label{fig:vis-point-maze-heatmap}
\vspace*{5mm}
\end{figure}

\begin{figure}[t]
     \centering
     \includegraphics[width=1.0\linewidth]{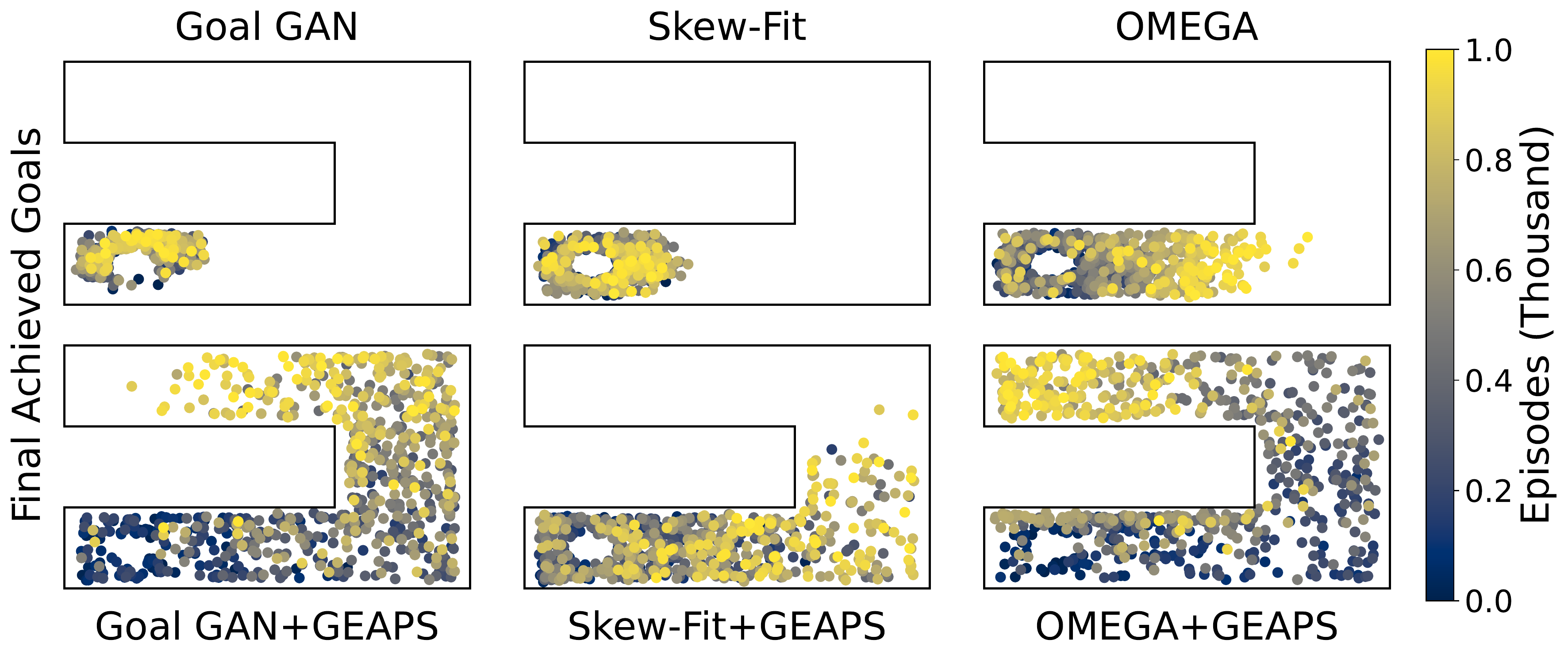}
     \vspace*{-5mm}
    \caption{Visualization of the final achieved goals  in \texttt{AntMaze}: the baselines (top) vs. the augmented models (bottom), where the training evolution process is indicated with the heatmap. }
    \label{fig:vis-ant-maze-heatmap}
%\vspace*{5mm}
\end{figure}

As shown in the top row of Figure~\ref{fig:vis-point-maze-heatmap}, all three baselines cannot reach the entire area of  \texttt{PointMaze} at the end of 1,600 episodes. Most goals reached by Goal GAN are located in the left half of the maze near the starting location. Most goals reached by Skew-Fit are located in a smaller area in the left half of the maze and goals mainly get stuck in two small areas close to or having a moderate distance to the starting location. OMEGA performs much better than other baselines as it covers the entire maze except those goals from the desired goal distribution in the top right corner. In contrast, it is evident from the bottom row of Figure~\ref{fig:vis-point-maze-heatmap} that our GEAPS makes all baselines cover a larger area and alleviates the so-called ``rich get richer" problem by sampling goals uniformly towards covering the entire maze. In particular, our GEAPS helps the baselines reach goals that spread outward from easy to hard goals as training advances and enables OMEGA to quickly transition to goals in the desired goal area.

It is observed from Figure~\ref{fig:vis-ant-maze-heatmap} that by incorporating our GEAPS into three baselines, their behavioral changes on \texttt{AntMaze}  are similar to those on \texttt{PointMaze}. At the end of 1,000 episodes, no baselines can reach goals beyond the bottom of the hallway, and goals reached by Goal GAN and Skew-Fit are even trapped in the small areas close to the starting
location. In contrast, the augmented models take advantage of our GEAPS, hence are able to reach goals in much larger areas. It is evident from the bottom row of Figure~\ref{fig:vis-ant-maze-heatmap} that OMEGA+GEAPS has already explored the goals within the desired goal area and Goal GAN+GEAPS has reached quite close to this area.

\begin{figure}[t]
     \centering
     \includegraphics[width=1.0\linewidth]{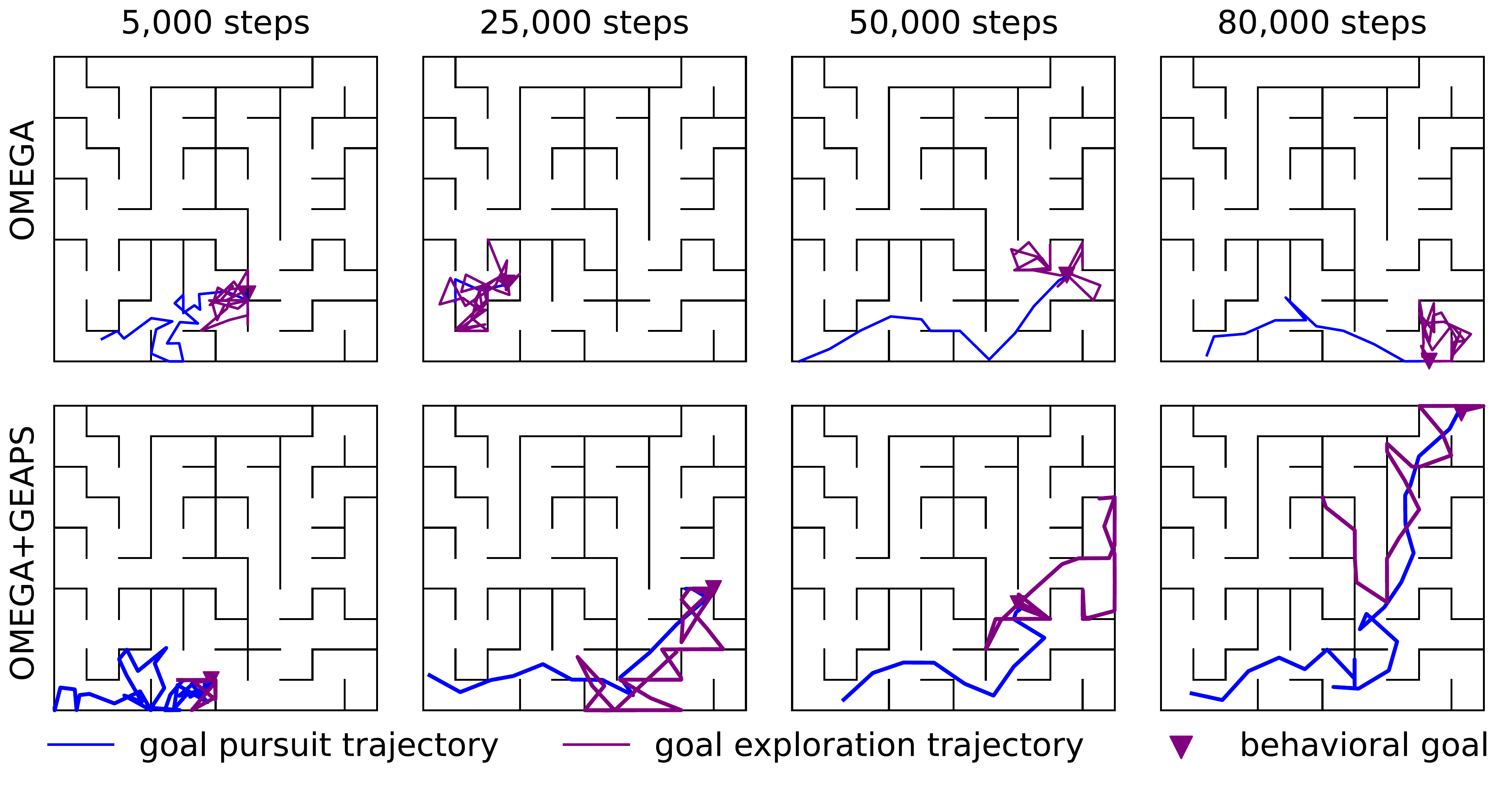}
     %\vspace*{-5mm}
    \caption{Visualization of the goal pursuit and goal exploration trajectories made by OMEGA (top) and OMEGA+GEAPS (bottom) at different training steps for \texttt{PointMaze}. }
    \label{fig:vis-point-maze-traj}
%\vspace*{-5mm}
\end{figure}

\begin{figure}[t]
     \centering
     \includegraphics[width=1.0\linewidth]{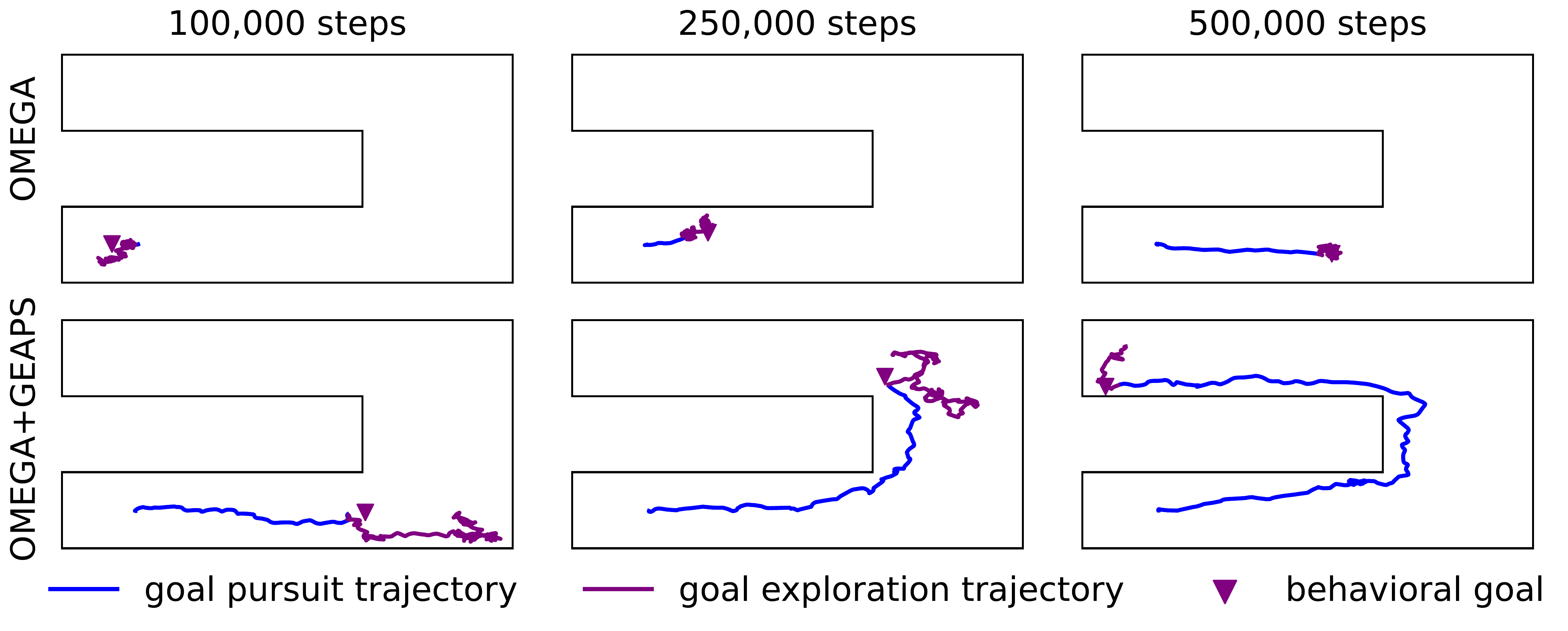}
     %\vspace*{-5mm}
    \caption{Visualization of the goal pursuit and goal exploration trajectories made by OMEGA (top) and OMEGA+GEAPS (bottom) at different training steps for  \texttt{AntMaze}. }
    \label{fig:vis-ant-maze-traj}
%\vspace*{-5mm}
\end{figure}

As OMEGA is the best performer among the three baselines and also uses a Go-Explore \cite{ecoffet2019go} style strategy for exploration, we further visualize the goal pursuit and exploration trajectories made by OMEGA and OMEGA+GEAPS at different training steps. As described in Section~\ref{sect:preliminry}, reaching a behavioral goal in goal pursuit triggers goal exploration. A trajectory can intuitively exhibit goal transitioned at different training steps to allow us to better understand the behavioral change resulting from our GEAPS.
As shown in Figures  \ref{fig:vis-point-maze-traj} and \ref{fig:vis-ant-maze-traj}, OMEGA generally explores goals close to the reached behavioral goals ``conservatively"  with a heuristic  \cite{pitis2020maximum} in goal exploration, while OMEGA+GEAPS explores goals in a larger area around the reached behavioral goals ``aggressively" by means of the frequently occurring goal-transition patterns encoded in the pre-trained skills, which vividly demonstrates the advantage of our proposed method in improving the exploration effectiveness during learning.

\begin{sidewaysfigure*}[htbp]
\begin{minipage}{\linewidth}
     \centering
     \includegraphics[width=1.0\linewidth]{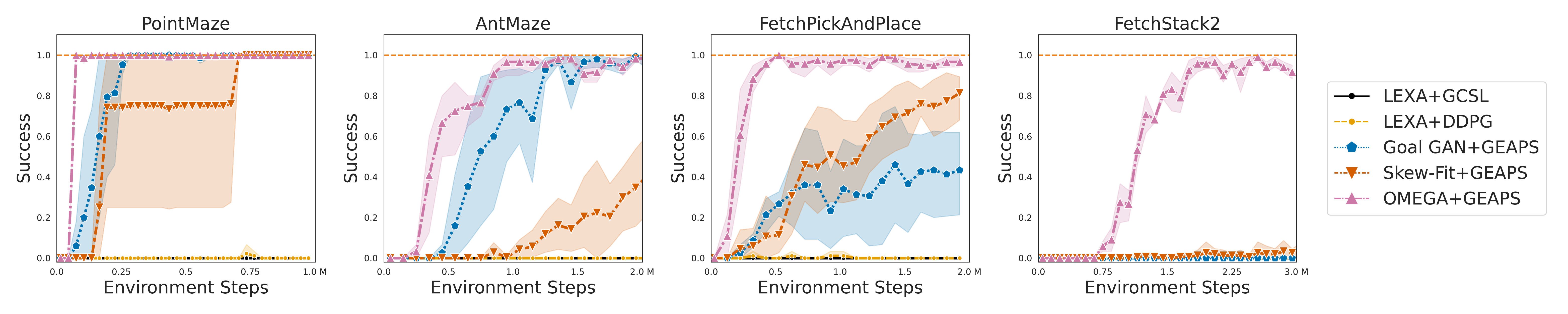}
    \caption{Test success on the desired goal distribution throughout training on four environments for our augmented models and LEXA explorer-based models. }
    \label{fig:vis-comparative-sota}
\end{minipage}
\begin{minipage}{\linewidth}
     \vspace*{15mm}
     \centering
     \includegraphics[width=1.0\linewidth]{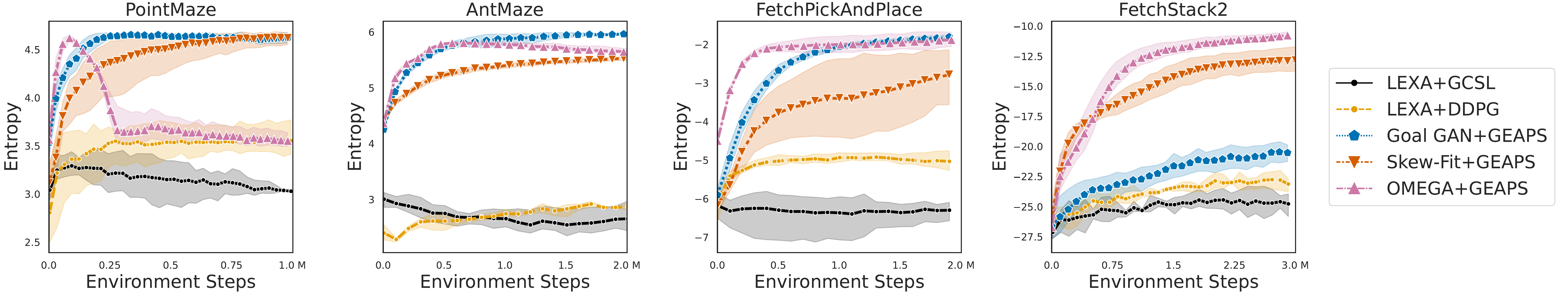}
    \caption{Empirical entropy of the achieved goal distribution throughout training for our augmented models and LEXA explorer-based models.}
    \label{fig:vis-entropy-sota}
\end{minipage}
\end{sidewaysfigure*}

\subsubsection{Comparison to LEXA Explorer}

\begin{figure}[t]
     \centering
     \includegraphics[width=0.7\linewidth]{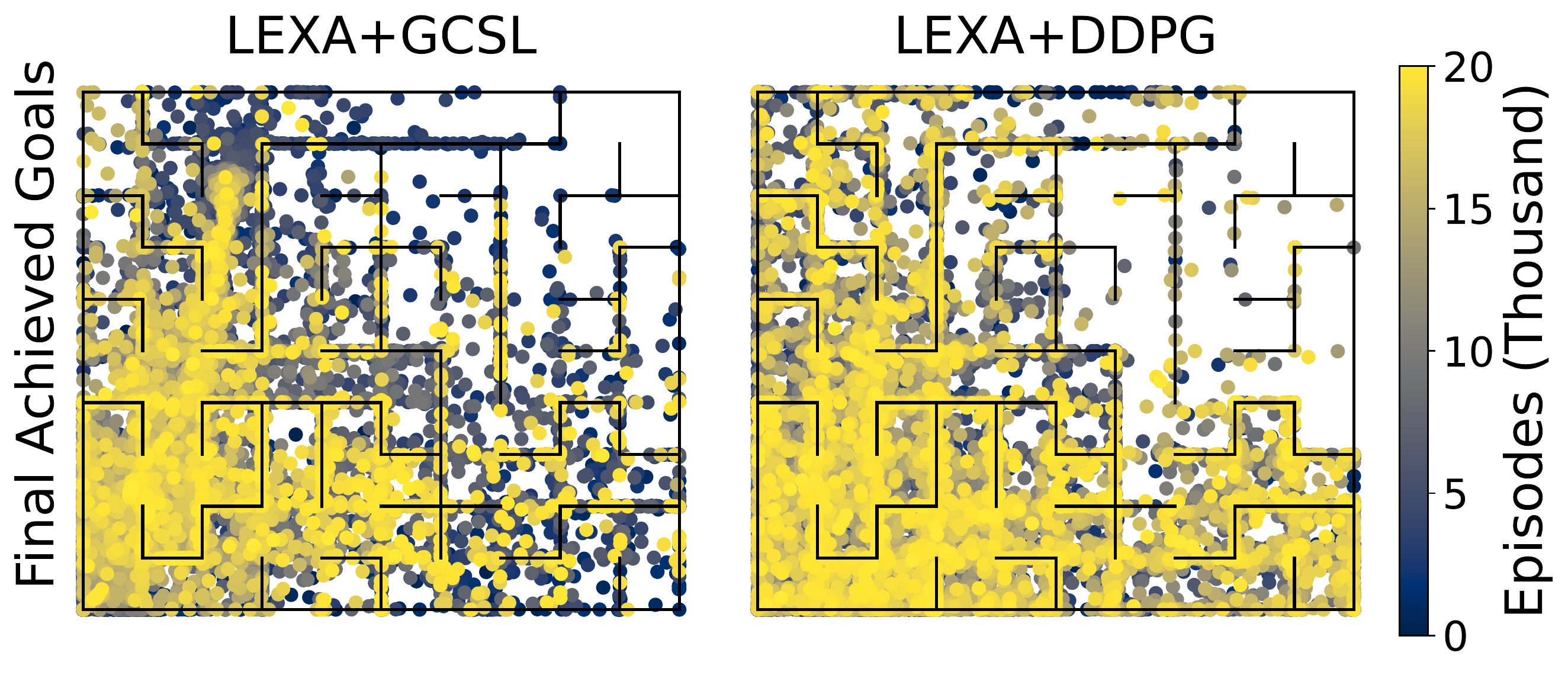}
    % \vspace*{-5mm}
    \caption{Visualization of the final achieved goals  on \texttt{PointMaze}: LEXA+GCSL and LEXA+DDPG, where the training evolution process is indicated by the heatmap.}
    \label{fig:vis-lexa-pointmaze}
    % \vspace*{-5mm}
    \end{figure}

To answer the third question, we compare the state-of-the-art  LEXA explorer-based goal exploration argumentation to our augmented models. As shown in Figure~\ref{fig:vis-comparative-sota}, within the same training budget, we only observe up to 7\% success rates on \texttt{PointMaze} with LEXA+DDPG and no success achieved by the LEXA explorer-based models on other experiments. As shown in Figure~\ref{fig:vis-entropy-sota}, the entropy of the LEXA explorer-based models are far below that of our augmented models.
The reasons may be two-fold: a) The LEXA explorer performs exploration via the disagreement of an ensemble of one-step world models and the disagreement is based on the novelty of states. Except for the \texttt{PointMaze}, the state space is not equivalent to the goal space and exploring more states does not necessarily contribute to exploring more goals. b) LEXA performs with the goal-pursuit behavior on those goals uniformly sampled from the replay buffer only, which prevents it from enhancing the experience around the explored goals. Form Figure~\ref{fig:vis-lexa-pointmaze}, we observe that the LEXA explorer is able to explore those areas near the desired goal distribution  on \texttt{PointMaze} while it fails to keep exploring those areas later. In contrast, our augmented models use different strategies to select those novel sub-goals to pursue, which enhances the experience around those novel goals.  Our augmented models prioritize pursuing those novel goals to broaden the relevant experience in the replay buffer.

In addition, training a world model in LEXA is time-consuming and requires abundant data, while our GEAPS only needs the skills pre-trained with our alternative learning objective. Those skills acquired by pre-training are applicable to any relevant downstream tasks. In summary,
the above results demonstrates that our model-free GEAPS is highly competitive with the SOTA model-based explorer in LEXA especially for the tasks of which state space is not equivalent to their goal space.

\subsubsection{Results on Skill Learning}
\label{subsect:skill-comparison}
\begin{figure}[htbp]
     \centering
     \begin{subfigure}[t]{0.42\textwidth}
         \includegraphics[width=\textwidth]{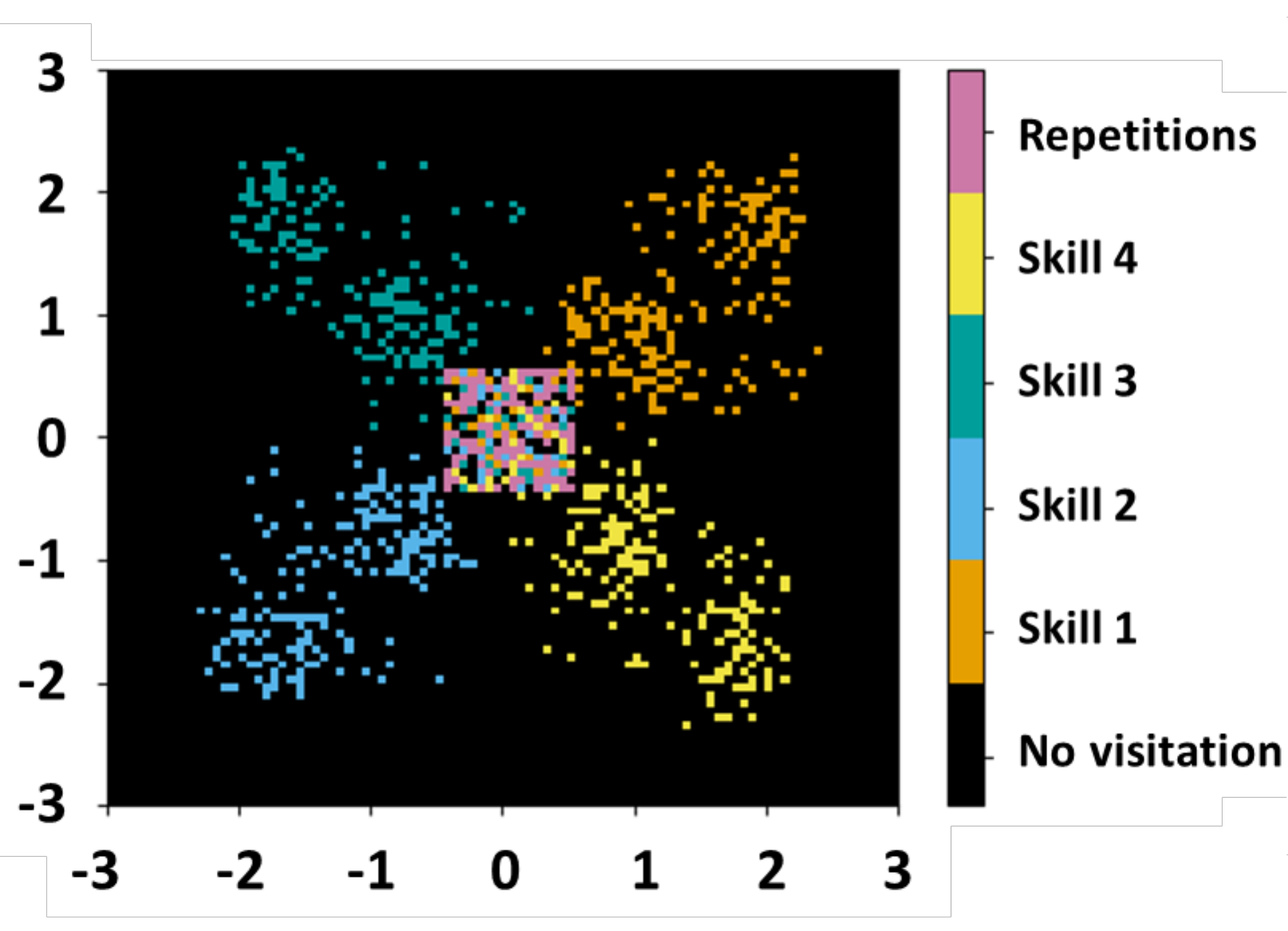}
         \caption{}
     \end{subfigure}
     \begin{subfigure}[t]{0.42\textwidth}
         \includegraphics[width=\textwidth]{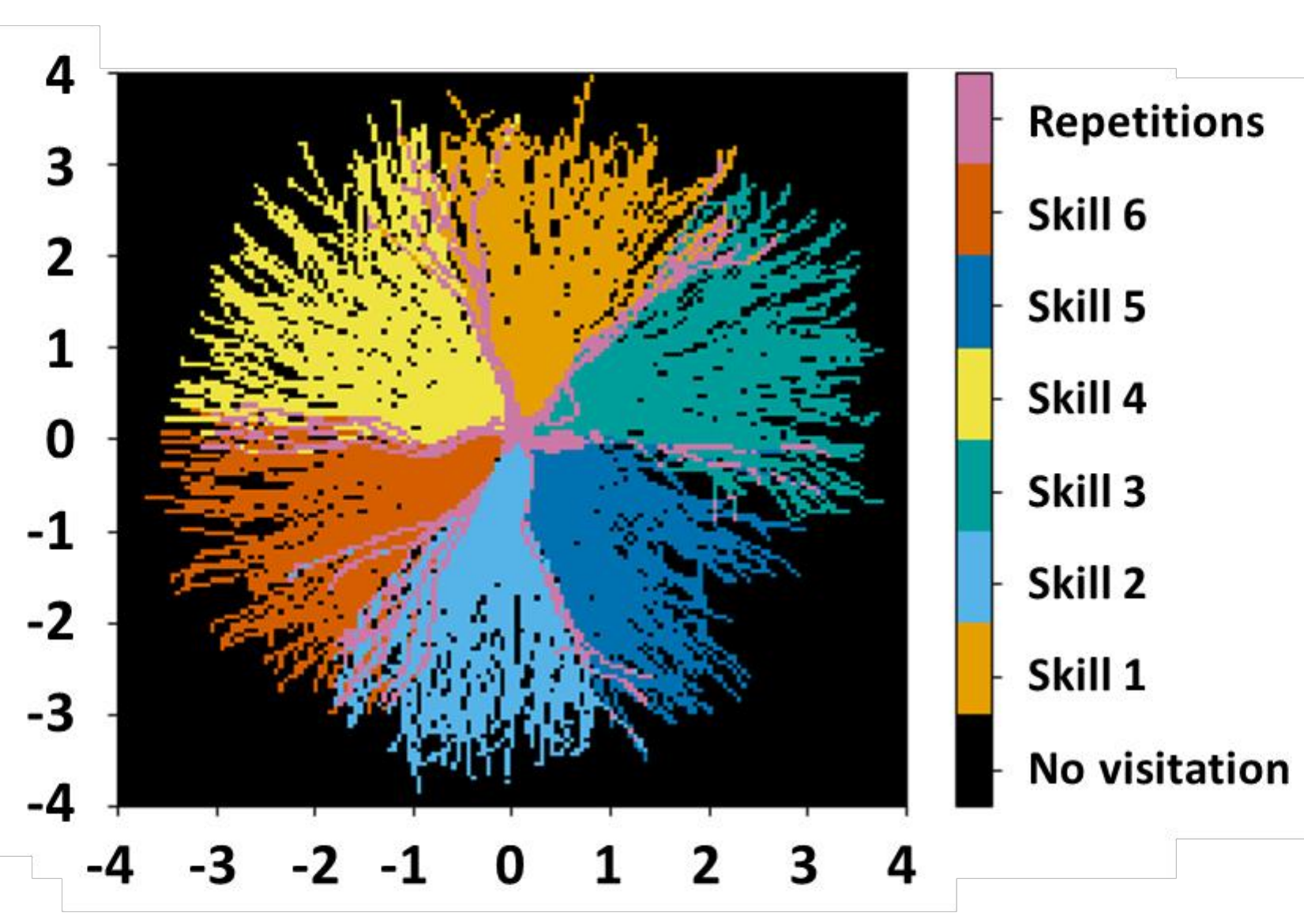}
         \caption{}
     \end{subfigure}
    \caption{Trajectories of pre-trained skills acquired by our skill learning method. (a) \texttt{PointMaze} in an empty maze. (b) \texttt{Ant}  in an empty maze.}
    \label{fig:vis-pretrain-skills}
\end{figure}

To answer the fourth question, we first visualize the pre-trained skills resulting from our learning objective presented in Sections \ref{subsect:skill-learn}.
Figure~\ref{fig:vis-pretrain-skills} illustrates the trajectories of pre-trained skills for \texttt{PointMaze} and \texttt{AntMaze}. In Figure~\ref{fig:vis-pretrain-skills}(a), we plot 50 trajectories for each skill pre-trained for \texttt{PointMaze} in an empty $5\! \times \!5$ maze. It is evident that the learned skills  guide the agent to navigate along different diagonal directions so that the agent can transit to another grid quickly. The skills are intuitive and their effectiveness in boosting the sample efficiency have been demonstrated by the results reported in \ref{subsect:augment} and \ref{subsect:visual}.
Figure~\ref{fig:vis-pretrain-skills}(b) shows the trajectories of the robot with the skills on the pre-training environment for \texttt{AntMaze} with equal probability for 100 episodes. As seen in Figure~\ref{fig:vis-pretrain-skills}(b), such skills have good coverage of goals in almost all directions. Besides, each skill evenly covers almost the same portion and no skill predominates the coverage of goals.

\begin{figure}
     % \vspace*{-9mm}
     \centering
     \includegraphics[width=0.85\linewidth]{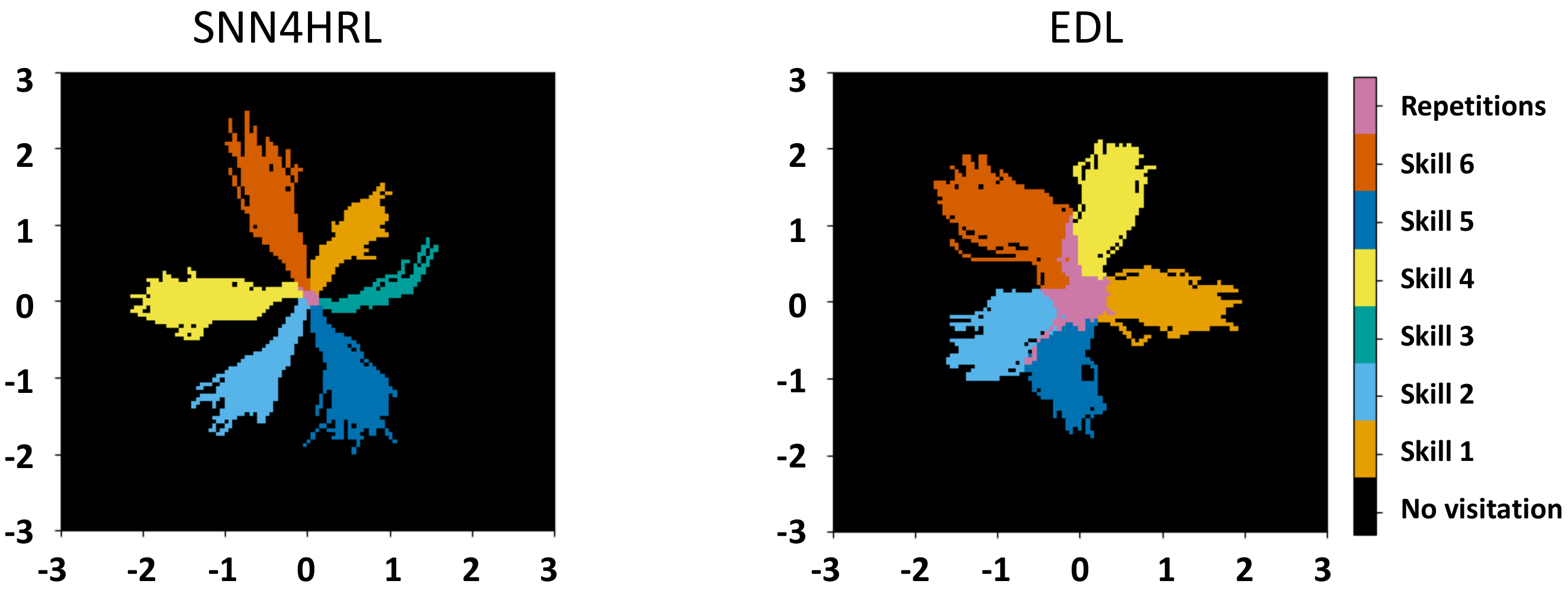}
      % \vspace*{-5mm}
    \caption{Trajectories of pre-trained skills learned by SNN4HRL and EDL for \texttt{AntMaze}.}
    \label{fig:vis-skill-traj}
\end{figure}
\noindent
\begin{figure}[htbp]
     \centering%
     \begin{subfigure}[t]{0.265\textwidth}
     \includegraphics[width=\textwidth]{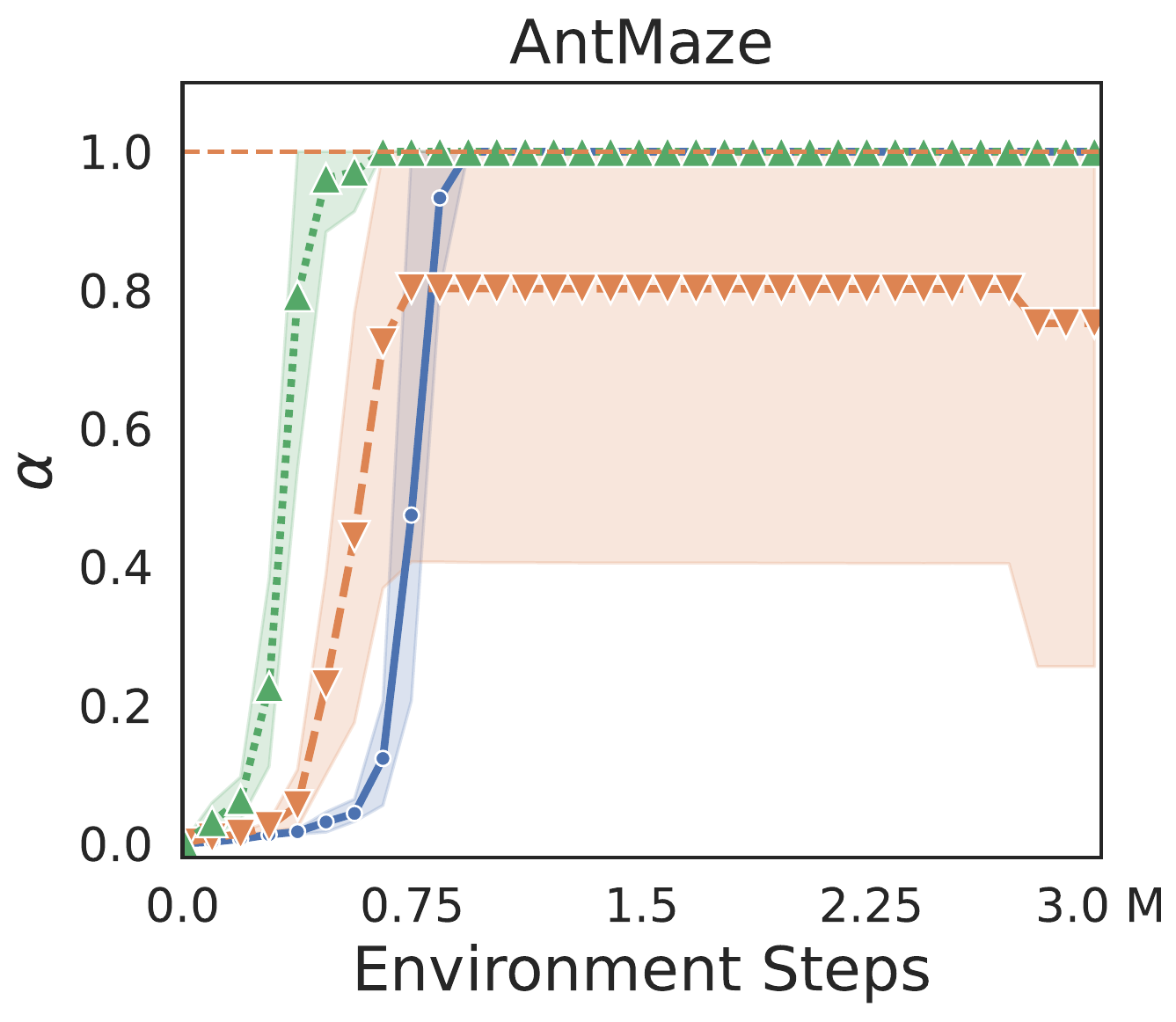}
         \caption{}
         %\label{fig:vis-skill-alpha}
     \end{subfigure}%
     \hfill
     \begin{subfigure}[t]{0.65\textwidth}
         \includegraphics[width=\textwidth]{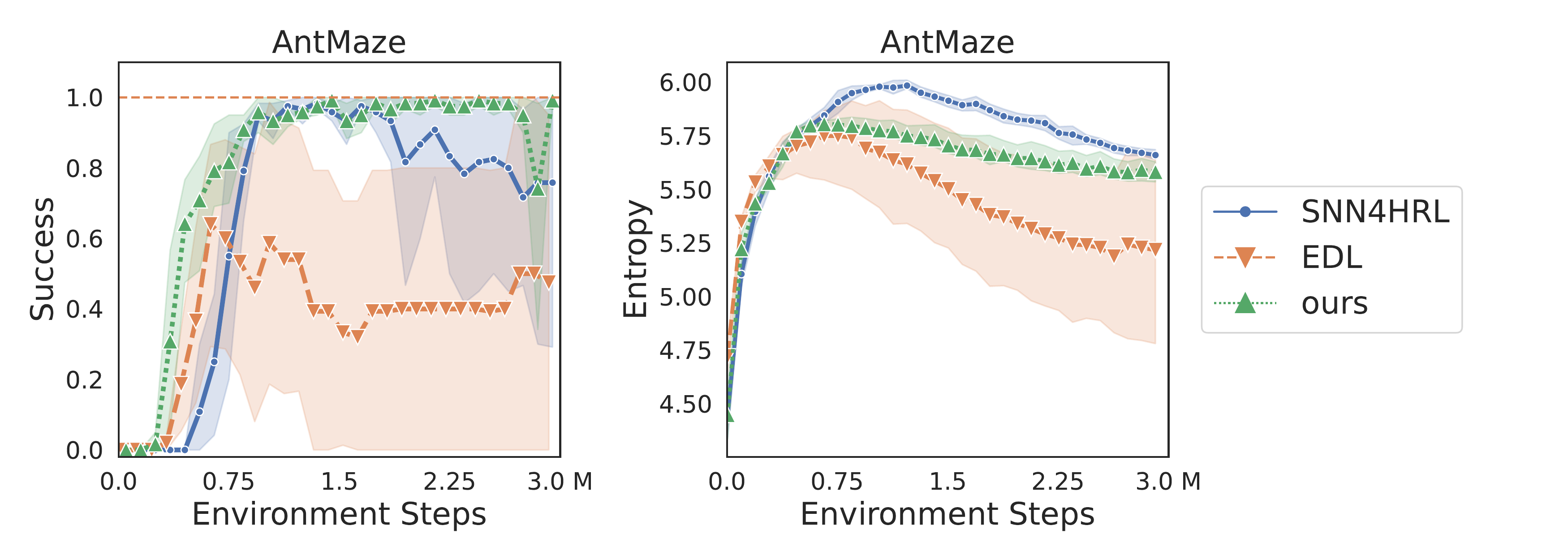}
         \caption{}
         %\label{fig:vis-skill-succ-entropy}
    \end{subfigure}%
\caption{(a) The dynamic weight $\alpha$ in OMEGA to trade-off the distributions of achieved goals and desired goals in the distributions of sub-goals throughout training on \texttt{AntMaze}. (b) Test success on the desired goal distribution and empirical entropy of the achieved goal distribution on \texttt{AntMaze} for OMEGA+GEAPS with the pre-trained skills resulting from different skill learning methods.}
\label{fig:two-figs}
\end{figure}

For comparison, we further illustrate the trajectories of pre-trained skills by SNN4HRL \cite{florensa2017stochastic} and EDL \cite{campos2020explore} on the same pre-training environment for \texttt{AntMaze} in Figure~\ref{fig:vis-skill-traj}. In contrast to the skills acquired by our method in Figure~\ref{fig:vis-pretrain-skills}(b), it is evident from Figure~\ref{fig:vis-skill-traj} that both SNN4HRL and EDL cover much smaller areas and leave numerous directions uncovered. In our experiment, we observe that the skills acquired by SNN4HRL appear unstable, highly depending on the random seeds.

To investigate the impact of pre-trained skills in our GEAPS,  we employ the skills acquired by the different skill learning methods in OMEGA+GEAPS for performance evaluation on \texttt{AntMaze}.
In OMEGA \cite{pitis2020maximum}, the factor $\alpha$ calculated via Eq.~\ref{eq:omega-alpha-calculation} in Appendix \ref{appendix-1} is inversely proportional to the KL divergence between the distribution of achieved goals $p_{ag}$ and desired goals $p_{dg}$ and its value is capped at one. It serves as a dynamic weight to balance both distributions in a mixture distribution for sub-goal sampling and recalculated at the end of each episode.  When the $\alpha$ reaches one, the agent only samples sub-goals from the desired goal distribution, which marks the end of exploration about sub-goals other than desired goals. It is evident from Figure~\ref{fig:two-figs}(a) 
%\ref{fig:vis-skill-alpha} 
that the pre-trained skills by our method allow for reaching $\alpha=1$ around 0.7 million steps, while the SNN4HRL skills have to take around one million steps and the EDL skills never lead to  $\alpha=1$. It is further observed from Figure  \ref{fig:two-figs}(b) 
that our skill learning objective results in earlier success on \texttt{AntMaze}; i.e., the agent with our skills starts to explore the desired goal distribution within 0.2 million steps, while the agents with the  SNN4HRL and the EDL skills have to take around 0.5 million episodes and 0.3 million steps, respectively. In the later training stage, the agent with our skills maintains high success rates regardless of different random seeds. In contrast, the performance of the agents with the SNN4HRL and EDL skills is degraded substantially. Moreover, we observe that the agent with the EDL skills always fails to solve the \texttt{AntMaze} task. In summary, the above results suggest that our skill learning objective yields the quality skills required by our GEAPS.

\section{Discussion}
\label{sect:disc}

In this section, we discuss the limitations/issues arising from our work and make a connection between our method and other related works.

While the advantages of our approach have been demonstrated, several limitations and open problems still remain.
First, our approach relies on the pre-trained skills obtained by skill learning in the environments similar to a target task. Our approach will not work if such environments are unavailable.
It is also worth stating that the skill learning incurs an additional computational overhead but is rewarded with great exploration efficiency in GCRL to accomplish a sparse-reward long-horizon task.
Next, our theoretical analysis establishes the theoretical justification for the benefits of utilizing pre-trained skills and the effectiveness achieved through our exploration strategy under specific conditions. However, further theoretical analyses concerning broader conditions are still pending.
Then, the environments used for evaluation have pre-defined yet well-behaved goal spaces and goals have to be in a vectorial form. It is unclear on whether our approach works in the same manner for various scenarios, e.g., an agent has to specify and model/learn its own goal space \cite{pong2019skew}, and goals are in other forms \cite{liuetal2022gcrl-survey}, e.g., image and language goals.
After that, all the baselines used in our experiments are sub-goal selection based GCRL algorithms \cite{liuetal2022gcrl-survey}. Without a considerable extra effort, our GEAPS method cannot be applied to other types of GCRL algorithms such as the optimization-based and the relabelling GCRL algorithms \cite{liuetal2022gcrl-survey} for goal exploration augmentation.
Finally, our approach is memoryless and thus treats both achieved and new goals to be explored equally during data rollout. Equipped with a memory mechanism, our approach would prevent any visited states from being revisited to further improve the sampling efficiency. With memory and proper pre-trained skills, an agent may accomplish new tasks via searching without any further learning.

It is well known that skills and options have been used in hierarchical reinforcement learning (HRL) for for exploration and task simplification \cite{sutton1998between}. However, in the context of GCRL, the direct applicability of pre-trained skills for goal attainment and maintenance is quite limited. This is due to the potential for overshooting goals or stochastic reaching, as well as the narrow focus of skills on specific goals \cite{gehring2021hierarchical}. In contrast, our GEAPS method combines the benefits of pre-trained skills with the precision of primitive actions, aiming to enhance goal exploration and achieve goals effectively.
Below, we summarize several key distinctions between our GEAPS method and existing works that utilize skills/options for exploration in HRL. 
First, our GEAPS expands the utilization of entropy maximization as a new learning objective in GCRL. By optimizing both achieved and prospective goals, our GEAPS enhances the efficiency of goal exploration. We specifically emphasize goal exploration and incorporate goal-transition patterns into the learning process, enabling more effective exploration even in the absence of precise dynamic knowledge. To the best of our knowledge, these distinctive features cannot be found in existing works on HRL in the context of GCRL.
Next, in HRL, a higher-level agent selects from these options, treating them as indivisible actions or atomic actions. Despite exploring goals while executing a skill, HRL often necessitates revisiting goals using more granular options.  In contrast, the skills trained in our GEAPS maximize their exploration capabilities based on goal-transition patterns specific to GCRL, allowing for interactions with a broader array of goals during execution. Our method enhances the efficiency of goal exploration and distinguishes our work from conventional HRL practices that prioritize re-engaging with the same set of goals. Even if the skills are pre-trained as sub-policies for specific sub-tasks in HRL \cite{gehring2021hierarchical}, each skill tends to primarily focus on a single goal associated with one of the sub-tasks. During execution, this narrow focus can severely limit the skill's ability to interact with a much wider range of goals that arise in GCRL.
Then, distinct from the HRL approach, which typically presumes task decomposition through options, our method does not mandate the completion of tasks strictly through pre-trained skills. Rather, within the context of our GEAPS, these skills are intentionally trained to enhance their efficacy in goal exploration, drawing upon goal-transition patterns particular to GCRL. Pre-trained skills, developed with a focus on these specific goal-transition patterns, empower our GEAPS to foster efficient exploration. Our method aligns closely with the innate exploratory behaviors observed in humans and animals, thus encouraging more intuitive interactions with the environment.
Finally, we acknowledge theoretical analyses on the exploration benefits of skills and options in HRL, such as the UCRL-SMDP framework \cite{fruit2017exploration}  that provides rigorous regret bounds for MDPs with options. However, the direct transfer of UCRL-SMDP to GCRL poses challenges due to disparities in reward mechanisms and the lack of historical data for novel goals. In contrast, our GEAPS addresses these challenges by efficiently navigating exploration in the absence of precise dynamic knowledge. While UCRL-SMDP may not directly aid in exploring unknown areas, a big challenge encountered in our work, it holds promise for enhancing policy optimization to efficiently reach already explored goals in the goal pursuit stage within the generic GCRL framework.

\section{Conclusion}

In this paper, we have proposed a novel learning objective that optimizes the entropy of both achieved and new goals in sub-goal selection based goal-conditioned reinforcement learning (GCRL). By optimizing this objective, we enhance the efficiency of goal exploration in complex environments, ultimately improving the performance of GCRL algorithms.

Our method incorporates skill learning, where frequently occurring goal-transition patterns are mined and composed into skills. These pre-trained skills are then utilized in goal exploration, allowing the agent to efficiently discover novel sub-goals. Through extensive evaluation on various sparse-reward long-horizon benchmark tasks and a theoretical analysis, we have demonstrated that integrating our method into state-of-the-art GCRL baselines significantly enhances their exploration efficiency while maintaining or improving their performance. 
The results of our research highlight the importance of effective goal exploration in addressing the challenges of sparse-reward long-horizon tasks. By augmenting the sub-goal section of GCRL models with our model-free goal exploration method, we achieve better coverage of the state space and improve sampling efficiency.

In our future work, there are several avenues for further investigation. First, we plan to conduct further theoretical analyses concerning broader 
conditions to gain deeper insights into the properties and guarantees of our proposed method. This will provide a solid foundation for understanding its advantages, limitations and potential extensions. Additionally, we aim to explore the application of our method in domains with image data, where the state space is more complex and requires specialized techniques.

In conclusion, our work contributes to the advancement of goal-conditioned reinforcement learning by offering an efficient goal exploration augmentation method. We believe that our research opens up new possibilities for addressing challenging sparse-reward long-horizon tasks in complex environments.

%\newpage

\backmatter

%\newpage
\vspace*{5mm}

\begin{appendices}

\counterwithout{theorem}{section}

\section{Proof of Propositions}
\label{appendix-0}

In this appendix, we provide proofs for the propositions formulated in Section~\ref{Sect:method} of the main text.

%Proof for Proposition 1
\begin{proposition}
Let $H_{ag}'(\mathcal{G})$ represent the updated entropy of achieved goals following the goal exploration. This entropy is bounded from below by the sum of the weighted entropies of the original achieved goals and the goals encountered during goal exploration, namely, $c~H_{ag}(\mathcal{G})$ and $(1-c)~H_e(\mathcal{G})$. That is,
\begin{align}
    H_{ag}'(\mathcal{G}) &\geq cH_{ag}(\mathcal{G}) + (1-c)H_e(\mathcal{G}). \nonumber
    %\label{eq:proposition-1}
\end{align}
\end{proposition}
\begin{boldproof}
We commence from the entropy definition of $H_{ag}'(\mathcal{G})$:
\begin{align}
    H_{ag}'(\mathcal{G}) &= -(c~p_{ag}(\mathcal{G}) + (1-c)~p_e(\mathcal{G}))~\text{log}~(c~p_{ag}(\mathcal{G}) + (1-c)~p_e(\mathcal{G})). \nonumber
   % \label{eq:ineq_proof_1}
\end{align}
Applying Jensen's inequality due to the concave property of entropy, we find:
\begin{align}
    H_{ag}'(\mathcal{G}) &\geq -c~p_{ag}(\mathcal{G})~\text{log}~p_{ag}(\mathcal{G}) - (1-c)~p_e(\mathcal{G})~\text{log}~p_e(\mathcal{G}). \nonumber
    %\label{eq:ineq_proof_2}
\end{align}
Recognizing $-c~p_{ag}(\mathcal{G})~\text{log}~p_{ag}(\mathcal{G})$ as $cH_{ag}(\mathcal{G})$ and $-(1-c)~p_e(\mathcal{G})~\text{log}~p_e(\mathcal{G})$ as $(1-c)H_e(\mathcal{G})$, we thus establish:
\begin{align}
    H_{ag}'(\mathcal{G}) &\geq cH_{ag}(\mathcal{G}) + (1-c)H_e(\mathcal{G}). \nonumber
    %\label{eq:ineq_proof_3}
\end{align}
This completes the proof, demonstrating that the updated entropy $H_{ag}'(\mathcal{G})$ is bounded by the weighted sum of the original entropies. 
\end{boldproof}

%Proof for Proposition 2 
\begin{proposition}
    The optimal exploration policy leading to $\Omega^*$ with the distribution $p_{\Omega^*}$ can be composed via a set of skills $\mathcal{Z}$ ($|\mathcal{Z}|<<|\Omega^*|$).
\end{proposition}

\begin{boldproof}
    We can cluster $|\Omega^*|$ into $|\mathcal{Z}|$ clusters and each cluster is represented by a latent vector $\pmb z\sim\mathcal{Z}$. Then, we have the corresponding distributions related to $\pmb z$.
    \begin{align}
        p(\pmb z) &= \sum_{\tau\in\Omega^*} p_{\Omega^*}(\tau)\mathbf{1}(\tau\in \pmb z), \label{Eq:appendix2-1-1}\\
        p(\tau|\pmb z) &= \frac{p_{\Omega^*}(\tau)p(\pmb z|\tau)}{p(z)} = \frac{p_{\Omega^*}(\tau)\mathbf{1}(\tau\in \pmb z)}{p(\pmb z)}.
    \label{Eq:appendix2-1-2}
    \end{align}
    In the above expressions, $\mathbf{1}(\tau\in \pmb z)$ denotes the indicator function, which equals 1 if $\tau$ belongs to the cluster represented by $z$ and 0 otherwise. In this setting, we can transform $\hat{H}_e^*({\mathcal{G}})$ with Eqs.~\ref{Eq:appendix2-1-1} and \ref{Eq:appendix2-1-2} into
    \begin{equation}
        \hat{H}_e^*({\mathcal{G}}) = I(\mathcal{Z}; \mathcal{G}) + H(\mathcal{G|Z}). 
        \nonumber
        %\label{eq:optimal_entropy}
    \end{equation}
    Although the mutual information term, $I(\mathcal{Z}; \mathcal{G})$, may decrease, the conditional entropy term, $H(\mathcal{G|Z})$, increases, maintaining the sum unchanged. For generating the optimal trajectories within each cluster, we can train a skill to produce those trajectories. The total number of such skills is $|\mathcal{Z}|$ and the condition $|\mathcal{Z}|<<|\Omega^*|$ can be fulfilled with appropriate clustering. During exploration, each skill corresponding to $\pmb z\sim\mathcal{Z}$ is sampled with probability $p(\pmb z)$. In the execution of each skill, the trajectory $\tau$ is generated with probability $p(\tau|\pmb z)$.
\end{boldproof}

%Proof for Proposition 3
\begin{proposition} \label{Appendix:prop-3}
Given the horizon $T$, every trajectory $\tau$ can be decomposed into a sequence of goal-transition patterns.
\end{proposition}

\begin{boldproof}
Our proof initiates by deconstructing the trajectory $\tau$ into two distinct sequences: the state sequence $S_\tau=(s_i)_{i=0}^{T}$ and the action sequence $A_\tau=(a_i)_{i=0}^{T-1}$. Upon acquiring $S_\tau$, we derive the corresponding goal sequence $G_\tau=(\phi(s_i))_{i=0}^{T}$. This goal sequence is subsequently partitioned into its maximal homogeneous segments, each embodying repetitions of a singular unique goal. The number of such segments is denoted as $N_g(\tau)$. For each of these segments, we annotate the specific goal and the time step of its first occurrence, denoted as ${((g_i, t_i))}_{i=0}^{N_g(\tau)-1}$. Following this, we append the tuple $(\phi(s_T), T)$ to the sequence, resulting in ${((g_i, t_i))}_{i=0}^{N_g(\tau)}$. Consequently, the trajectory can be decomposed into a sequence of goal-transition patterns symbolized as $\{\psi_i\}_{i=0}^{N_g(\tau)-1}$, where each pattern $\psi_i$ is defined as
\begin{equation}
\psi_i=\{s_{t_i}^{agent}, s_{t_{i+1}}^{agent}, \Delta(g_{t_i}, g_{t_{i+1}}), (a_j)_{j=t_i}^{t_{i+1}-1}\}. \nonumber
\end{equation}
\end{boldproof}

%Proof for Proposition 4
\begin{proposition}
Given an exploration horizon of $T^e$, the substitution of
goal-transition patterns within each trajectory $\tau\in\Omega$ with
alternative patterns of smaller cardinality can yield equivalent
exploration outcomes using an average number of steps that is less than
or equal to the specified $T^e$.
\end{proposition}

\begin{boldproof}
For any trajectory $\tau\in\Omega$, it can be decomposed into a sequence
of goal-transition patterns $\{\psi_i\}_{i=0}^{N_g(\tau)-1}$ as outlined
in Proposition \ref{Appendix:prop-3}. There exists an alternative
goal-transition pattern to $\psi_i$ for the transition $\Delta(g_{t_i},
g_{t_{i+1}})$ as follows: $\forall \psi_i \in
\{\psi_i\}_{i=0}^{N_g(\tau)-1}, \exists \psi=\{s^{agent}_{start},
s^{agent}_{end}, \Delta G, \Delta A\} \in \Psi$, where $s^{agent}_{start}=
s_{t_i}^{agent}$, $s^{agent}_{end}=s^{agent}_{t_{i+1}}$, $\Delta G =
\Delta(g_{t_i}, g_{t_{i+1}})$ and $|\Delta A| \leq t_{i+1} - t_i$. By
substituting $\psi_i$ with the equivalent pattern necessitating the
fewest steps, we can derive a new sequence of goal-transition patterns
$\{\Tilde{\psi}_i\}_{i=0}^{N_g(\tau)-1}$ such that
$\sum_{i=0}^{N_g(\tau)-1}|\Tilde{\psi}_i|\leq T_e$.
\end{boldproof}

\section{Goal Exploration}
\label{appendix-1}

To facilitate the readability, we provide the further details omitted in Section 5 of the main text in this appendix, including the technical details of baselines and the state-of-the-art method LEXA explorer and  the implementation details of baselines and LEXA explorer used in our experiments.

\subsection{Technical Details}

\subsubsection{Goal GAN} 
Goal GAN \cite{florensa2018automatic} aims to select sub-goals of intermediate difficulties. Given the policy $\pi_k$ at iteration $k$ and a goal $g$, we denote its expected return as $R^g(\pi_k)$. Thus, the set of \textit{Goals of Intermediate Difficulty} (GOID) is defined as follows:
\begin{equation}
    {\rm GOID}_k \triangleq \{g: R_{min} \leq R^g(\pi_k)\leq R_{max}\},
\end{equation}
where $R_{min}$ and $R_{max}$ are the minimum and maximum expected return of goals for the agent to pursue, respectively. Also, $R_{min}$ and $R_{max}$ can be interpreted as the minimum and maximum success rates of reaching a goal within $T$ steps. 
%\lisheng{(NOTE: Shall we change the name of Goal GAN to Goaldisc as \cite{pitis2020maximum} because what I used is not actually a generative model.)} 
To identify the goals of intermediate difficulties, we adopt the same method used in \cite{pitis2020maximum}; i.e., a discriminator is trained to distinguish whether a behavioural goal can be achieved from a specific goal. During training, the start state and the behavioural goal of each trajectory are taken as input, and the binary target would be one only if the behavioural goal was achieved within the trajectory. During goal sampling, the initial state and goal candidates are fed to a trained discriminator as input, which predicts the success probability $R^g(\pi_k)$ of reaching each candidate. Based on the prediction, the agent can decide the GOID set to be sampled from. Then, the GOID is further ranked according to how far $R^g(\pi_k)$ is close to $0.5$.

\subsubsection{Skew-Fit} 
The key idea of Skew-Fit \cite{pong2019skew} is to increase the diversities of goals by maximising the entropy of achieved goals. Thus, Skew-Fit aims to train a generative model $q_\phi^\mathcal{G}$ that achieves maximum entropy on all the goals. To ensure its entropy is monotonically improved, it proposes to skew the distribution via sampling importance resampling  as follows:

\begin{align}
    p_{{skewed}_t}(g) \triangleq \frac{1}{Z_{\alpha_1}} q_{\phi_t}^\mathcal{G}(g)^{\alpha_1}\delta(g\in\mathcal{G}_\mathcal{B}),\label{eq:skew}\\
    Z_{\alpha_1} = \sum_{g\in\mathcal{G}_\mathcal{B}} q_{\phi_t}^\mathcal{G}(g)^{\alpha_1}(g), g \overset{\text{iid}}{\sim} p_{\phi_t}^\mathcal{G}. \label{eq:skew_norm}
\end{align}
    Here, $p_{\phi_t}^\mathcal{G}$ is the unknown underlying distribution of goals to be achieved via the policy at the $t$th iteration of training the generative model and is estimated via the approximation $p_{\phi_t}^\mathcal{G} \approx q_{\phi_t}^\mathcal{G}$. $\alpha_1$ (${\alpha_1}<0$) is used to balance the reliability of $q_{\phi_t}^\mathcal{G}(\mathcal{S})$ and the speed to increase the entropy of goal distribution. Then $q_{\phi_{t+1}}$ is trained to fit $p_{{skewed}_t}$, resulting in $q_{\phi_{t+1}} \approx p_{{skewed}_t}$. At the $(t+1)$th iteration, the goals can be sampled from $p_{{skewed}_t}$ or $q_{\phi_{t+1}}$.
%\lisheng{(NOTE: We also have hyper-parameters named as $\alpha$ and $\beta$ (used in $\beta$-VAE) in the main article, shall we rename them.)}

\subsubsection{OMEGA} 
Given a distribution of desired goals $p_{dg}$, OMEGA \cite{pitis2020maximum} aims at selecting a sub-goal that can minimize the KL divergence between $p_{dg}$ and the distribution of achieved goals $p_{ag}$.
\begin{equation}
  J_{original}(p_{ag}) = D_{KL}(p_{dg}||p_{ag}). \nonumber
\end{equation}
The above original learning objective is ill-conditioned and not finite for a long-horizon task since $p_{ag}$ and $p_{dg}$ do not overlap at the beginning. Therefore, this objective is amended via expanding the support of achieved goals to make $J_{original}(p_{ag})$ as soon as possible. It can be realized by the \textit{Maximum Entropy Goal Achievement} (MEGA) objective that maximizes the entropy of achieved goals as follows:
\begin{equation}
  J_{MEGA}(p_{ag}) = D_{KL}(\mathcal{U}({\rm supp}(p_{ag}))||p_{ag}). \nonumber
\end{equation}
where $\mathcal{U}({\rm supp}(p_{ag}))$ denotes the uniform distribution on the support of $p_{ag}$. Compared to MEGA, OMEGA uses a mixture distribution $p_{\alpha} = \alpha p_{dg} + (1-\alpha)\mathcal{U}({\rm supp}(p_{ag}))$ as the target in the optimization of KL divergence; i.e.,
\begin{equation}
  J_{OMEGA}(p_{\alpha}) = D_{KL}(p_{\alpha}||p_{ag}). \nonumber
\end{equation}
The way to achieve $\alpha$ suggested in \cite{pitis2020maximum} is as follows:
\begin{equation}
  \alpha = 1/\text{max}(b+D_{KL}(p_{dg} || p_{ag}), 1), 
  \label{eq:omega-alpha-calculation}
\end{equation}
where $b\leq 1$. To optimize the OMEGA objective, the agent would sample sub-goals from desired goals at $\alpha$-probability and achieved goals at (1-$\alpha$)-probability in the following way:
\begin{equation}
  \hat{g} = \text{arg min}_{\hat{g}\in\mathcal{B}}p_{ag}(\hat{g}).\label{eq:omega-min-density}
\end{equation}

\subsubsection{LEXA Explorer} LEXA \cite{mendonca2021discovering} is a model-based reinforcement learning algorithm with two components: explorer and  achiever. The explorer acts for active exploration and trained to explore curious states via a world model. The explorer is trained with unsupervised rewards based on the disagreements of an ensemble of 1-step transition models that predict the next world model states from a current model state. The ensemble of the one-step models can be expressed as
\begin{equation}
  \text{Ensemble: } f(s_t, \theta^m) = \hat{z}^m_{t+1}, m =1\dots M, \nonumber
\end{equation}
where $\hat{z}^m_{t+1}$ indicates the next model state predicted by model $m$ in the ensemble of $M$ models. Assume that there are $D$ dimensions totally in the model state, the reward of state $s$ is the averaged variance of the states predicted by the ensemble model across all dimensions:
\begin{equation}
    r^e(s_t) = \frac{1}{D}\sum_{d=1}^D \text{Var}_{m}[f(s_t, \theta^m)]_d. \label{eq:lexa-reward}
\end{equation}
For the achiever, we used the rewards from the environment to replace the unsupervised rewards used in \cite{mendonca2021discovering} for fair comparison in exploration. The achiever in our experiments is trained via the standard GCSL \cite{ghosh2020learning} in the open-source code provided by the authors where DDPG is used in the baselines.

\subsection{Implementation Details}
\subsubsection{DDPG} All baselines are implemented on the basis of DDPG \cite{lillicrap2015continuous}. The details of relevant hyperparameters used in DDPG are listed in Table~\ref{table:ddpg-hyperparameters}. The training frequency varies over different tasks as reported in Table~\ref{table:env-dependent-hyperparamters}.

\begin{table}[h!]
\caption{Hyperparamters in DDPG.}
\centering
\begin{tabular}{c | c}
 \hline
 \textbf{Hyperparamter} & \textbf{Value} \\ [0.5ex]
 \hline\hline
batch size & 2000 \\
actor learning rate & 1e-3 \\
critic learning rate & 1e-3 \\
optimizer &  Adam \cite{kingma2014adam} \\
activation & GeLU \cite{hendrycks2016gaussian} \\
hidden layer sizes (actor and critic) & (512, 512, 512) \\
% relabelling strategies & ?\\
% density models training frequencies & ? \\
% optimisation frequency & ? expect to be different on different baselines ()\\
target network update proportion & 0.05 \\
target network update frequency & 40 steps \\
initial random data collection & 5000 steps \\
epsilon for random exploration & 0.1 \\
replay buffer size & 5000,000 \\
discount factor & 0.98 (0.99 for AntMaze) \\  [1ex]
 \hline
\end{tabular}
\label{table:ddpg-hyperparameters}
\end{table}

\begin{table}[h!]
\caption{Hyperparamters for Different Tasks.}
\centering
\begin{tabular}{c | c | c}
 \hline
\textbf{ Environment} & \textbf{Hyperparameter} & \textbf{Value} \\ [0.5ex]
 \hline\hline
PointMaze & relabelling strategy & rfaab\_1\_4\_3\_1\_1 \\
  & train every & 1 \\
\hline
AntMaze & relabelling strategy & rfaab\_1\_4\_3\_1\_1 \\
 & train every & 1 \\
 \hline
FetchPickAndPlace & relabelling strategy & rfaab\_1\_5\_2\_1\_1 \\
& train every & 4 \\
\hline
FetchStack2 & relabelling strategy & rfaab\_1\_5\_2\_1\_1 \\
& train every & 10 \\[1ex]
 \hline
\end{tabular}
\label{table:env-dependent-hyperparamters}
\end{table}

\subsubsection{Relabelling Techniques} During training, we adopt the same relabelling strategies $\texttt{rfaab}$ used in \cite{pitis2020maximum}: mixing different relabelling techniques \texttt{real}, \texttt{future}, \texttt{actual}, \texttt{achieved} and \texttt{behavioral} at a fixed ratio. \texttt{Real} stands for no relabelling. \texttt{Future}, \texttt{actual}, \texttt{achieved}, \texttt{behavioral} indicate relabelling with goals from future achieved goals in the belonging trajectories, all historically desired goals, all historically achieved goals and all historically behavioural goals, respectively. Their relative ratios are used to specify the specific  technique. For example, \texttt{rfaab\_1\_4\_3\_1\_1} denote no relabelling on 10\% data and relabelling 40\% with \texttt{future}, 30\% with \texttt{achieved}, 10\% with \texttt{actual} goals and 10\% with \texttt{behavioral}. The relabelling strategies vary in different environments (see Table~\ref{table:env-dependent-hyperparamters} for details).

\subsubsection{Goal GAN} The neural network used as the discriminator has the same architecture as that of  the critic in DDPG  except that the sigmoid activation is used in the output layer. The discriminator is trained with a batch of 100 trajectories sampled from the 200 most recent ones for every 250 steps. The $R_{min}$ and $R_{max}$ are set to 0.25 and 0.75, respectively.

\subsubsection{SkewFit} 
Following the same settings in Skew-Fit \cite{pong2019skew}, we empoly the $\beta$-VAE as the generative model. Both the encoder and decoder of $\beta$-VAE have two hidden layers with [400, 300] ReLU units. Its latent dimension size is set to be the same as the size of the goal in the environment. In $\beta$-VAE, we set $\beta=10$ as 10 and the $\alpha_1=2.5$ in Eqs.~\ref{eq:skew} and \ref{eq:skew_norm}. We set the batch size as 64 for training $\beta$-VAE and 
adopt the same training setting in Skew-Fit \cite{pong2019skew}: training every 4,000 steps for 1000 batches in the first 40,000 steps and every 4,000 steps for 200 batches afterwards.

\subsubsection{OMEGA} 
We adopt the same settings used in OMEGA \cite{pitis2020maximum} as follows. We set $b$ in Eq.~\ref{eq:omega-alpha-calculation} to be -3.0. To approximate the probability $p_{ag}(\hat{g})$ for a given $\hat{g}$ in Eq.~\ref{eq:omega-min-density}, we use the kernel density estimator (KDE) \cite{rosenblatt1956remarks} with 0.1 bandwidth and Gaussian kernel as our density model. We fit the KDE model to 10,000 normalized achieved goals sampled from the replay buffer for every optimization step.

\subsubsection{LEXA} We adopt RSSM \cite{hafner2019learning} as the world model. There are three hidden layers with [128, 128, 64] with [400, 300] ReLU units in both the encoder and the decoder. The hidden layer size for the recurrent model is set to 128. The sizes of the deterministic state and stochastic state are 128 and 32, respectively. We use 10 one-step world models (i.e., $M$=10) to construct an ensemble world model that calculates the exploration rewards specified in Eq.~\ref{eq:lexa-reward}. Each component world model consists of four hidden layers where each hidden layer has 400 ELU units  \cite{clevert2015fast}. In the GCSL implementation, we use the same actor architecture and the same learning rate used in DDPG as shown in Table~\ref{table:ddpg-hyperparameters} where only the \texttt{future} relabelling techniques are used during training.

\section{Skill Learning}
\label{appendix-2}

In this section, we  provide the information on the main hyper-parameters used in our comparative study in skill learning.

\subsection{SNN4HRL and Ours} 
The skill policy network used in SNN4HRL \cite{florensa2017stochastic} has two hidden layers of 64 Tanh units. The policy network is trained with TRPO \cite{schulman2015trust} with learning rate 0.01 and batch size 50,000 for 300 iterations. For the reward computation, we discretize the goal space into grids of size $0.2\times 0.2$ to calculate the rewards.
Our skill learning method shares the same hyper-parameters with SNN4HRL methods except for the entropy term $\mathcal{H(\mathcal{G}|\mathcal{Z})}$ weighted by 0.1.

\subsection{EDL} 
 EDL \cite{campos2020explore} consists of state marginal matching (SMM) \cite{lee2019efficient}, VQ-VAE \cite{van2017neural} and skill learning. We adopt the same hyper-parameters and learning methods used in the original square maze environments \cite{campos2020explore}. Nevertheless, to adapt it to the \texttt{Ant} environments, we increase the environment steps per cycle to 30 and batch size to 1,024 in SMM and set the number of epochs as 100 for VQ-VAE training and  the number of rollouts per cycle as 6. Finally, the training epochs for skill training is set to 10 .
 
%%=============================================%%
%% For submissions to Nature Portfolio Journals %%
%% please use the heading ``Extended Data''.   %%
%%=============================================%%

%%=============================================================%%
%% Sample for another appendix section                 %%
%%=============================================================%%

%% \section{Example of another appendix section}\label{secA2}%
%% Appendices may be used for helpful, supporting or essential material that would otherwise
%% clutter, break up or be distracting to the text. Appendices can consist of sections, figures,
%% tables and equations etc.

\end{appendices}

%%===========================================================================================%%
%% If you are submitting to one of the Nature Portfolio journals, using the eJP submission   %%
%% system, please include the references within the manuscript file itself. You may do this  %%
%% by copying the reference list from your .bbl file, paste it into the main manuscript .tex %%
%% file, and delete the associated \verb+\bibliography+ commands.                            %%
%%===========================================================================================%%

% \noindent
% If any of the sections are not relevant to your manuscript, please include the heading and write `Not applicable' for that section.

\vspace*{5mm}
\bibliography{uom-bibliography}% common bib file
%% if required, the content of .bbl file can be included here once bbl is generated
%%\input sn-article.bbl

%% Default %%
%%\input sn-sample-bib.tex%

\end{document}